\documentclass[DIV15, a4paper]{scrartcl}

\usepackage[ruled, linesnumbered, vlined, commentsnumbered]{algorithm2e}
\usepackage{algpseudocode}
\usepackage{authblk}
\usepackage{prettyref}
\usepackage{graphicx}
\usepackage{subfig}
\usepackage{booktabs}
\usepackage{colortbl}
\usepackage{amssymb}
\usepackage{amsfonts}
\usepackage[cmex10]{amsmath}
\usepackage{amsthm}
\usepackage{multirow}
\usepackage{tabularx}
\usepackage{footnote}
\usepackage{threeparttable}
\usepackage{hyperref}
\usepackage{amsopn}
\usepackage{hhline}
\usepackage{cite}
\usepackage{subfig}
\usepackage{setspace}
\usepackage{epstopdf}
\usepackage{helvet}
\usepackage[eulergreek]{sansmath}
\usepackage{textcomp}
\usepackage{array}

\usepackage{tikz}
\usepackage{pgfplots}
\usepackage{pgfplotstable}

\usepackage{xcolor}  % Required for custom colors
\definecolor{myblue}{RGB}{34,31,217}

\definecolor{mycyan}{gray}{.7}

\newcommand{\pref}{\prettyref}

\newrefformat{fig}{Fig.~\ref{#1}}
\newrefformat{tab}{Table~\ref{#1}}
\newrefformat{sec}{Section~\ref{#1}}
\newrefformat{app}{Appendix~\ref{#1}}
\newrefformat{alg}{Algorithm~\ref{#1}}
\newrefformat{property}{Property~\ref{#1}}
\newrefformat{theorem}{Theorem~\ref{#1}}
\newrefformat{corollary}{Corollary~\ref{#1}}
\newrefformat{proposition}{Proposition~\ref{#1}}
\newrefformat{def}{Definition~\ref{#1}}
\newrefformat{eq}{equation~(\ref{#1})}

\usepackage{graphicx}
\definecolor{Gray}{gray}{0.9}
\usepackage{scrpage2}

\usepackage{lscape}

\begin{document}

\title{\textbf\LARGE\fontfamily{cmss}\selectfont Matching-Based Selection with Incomplete Lists for Decomposition Multi-Objective Optimization\footnote{This paper has been accepted for publication by IEEE Transactions on Evolutionary Computation. Copyright will be transferred to IEEE Press without notice.}}

%\author[1]{\normalsize\fontfamily{lmss}\selectfont Ke Li\footnote{Department of Computer Science, University of Exeter, EX4 4QF, UK. Email: k.li@exeter.ac.uk}\hspace{10mm} Kalyanmoy Deb\footnote{COIN Laboratory, Department of Electrical and Computer Engineering, Michigan State University, MI 48824, USA. Email: kdeb@egr.msu.edu}\hspace{10mm} Xin Yao\footnote{CERCIA, School of Computer Science, University of Birmingham, B15 2TT, UK. Email: x.yao@cs.bham.ac.uk}}
\author[1]{\normalsize\fontfamily{lmss}\selectfont Mengyuan Wu}
\author[2]{\normalsize\fontfamily{lmss}\selectfont Ke Li}
\author[1]{\normalsize\fontfamily{lmss}\selectfont Sam Kwong}
\author[3]{\normalsize\fontfamily{lmss}\selectfont Yu Zhou}
\author[1]{\normalsize\fontfamily{lmss}\selectfont Qingfu Zhang} 
\affil[1]{\normalsize\fontfamily{lmss}\selectfont Department of Computer Science, City University of Hong Kong}
\affil[2]{\normalsize\fontfamily{lmss}\selectfont College of Engineering, Mathematics and Physical Sciences, University of Exeter}
\affil[3]{\normalsize\fontfamily{lmss}\selectfont College of Computer Science and Software Engineering, Shenzhen University}
\affil[$\ast$]{\normalsize\fontfamily{lmss}\selectfont Email: k.li@exeter.ac.uk, \{mengyuan.wu, yzhou57-c\}@my.cityu.edu.hk, \{cssamk, qingfu.zhang\}@cityu.edu.hk}

\renewcommand\Authands{ and }

\date{}
\maketitle

{\normalsize\fontfamily{lmss}\selectfont\textbf{Abstract: } }
The balance between convergence and diversity is the cornerstone of evolutionary multi-objective optimization. The recently proposed stable matching-based selection provides a new perspective to handle this balance under the framework of decomposition multi-objective optimization. In particular, the one-one stable matching between subproblems and solutions, which achieves an equilibrium between their mutual preferences, is claimed to strike a balance between convergence and diversity. However, the original stable marriage model has a high risk of matching a solution with an unfavorable subproblem, which finally leads to an imbalanced selection result. In this paper, we introduce the concept of incomplete preference lists into the stable matching model to remedy the loss of population diversity. In particular, each solution is only allowed to maintain a partial preference list consisting of its favorite subproblems. We implement two versions of stable matching-based selection mechanisms with incomplete preference lists: one achieves a two-level one-one matching and the other obtains a many-one matching. Furthermore, an adaptive mechanism is developed to automatically set the length of the incomplete preference list for each solution according to its local competitiveness. The effectiveness and competitiveness of our proposed methods are validated and compared with several state-of-the-art evolutionary multi-objective optimization algorithms on 62 benchmark problems.

{\normalsize\fontfamily{lmss}\selectfont\textbf{Keywords: } }
Convergence and diversity, stable matching with incomplete lists, adaptive mechanism, decomposition, multi-objective optimization. 

% !Tex Root = main.tex

\section{Introduction}
\label{sec:introduction}

The multi-objective optimization problem (MOP) considered in this paper is defined as follows~\cite{DebBook}:
\begin{equation}
\begin{array}{l}
    \mathrm{minimize} \quad \mathbf{F}(\mathbf{x})=(f_1(\mathbf{x}),\cdots,f_m(\mathbf{x}))^{T}\\
	\mathrm{subject\ to} \quad \mathbf{x}\in\Omega
\end{array}
\label{eq:MOP}
\end{equation}
where $\mathbf{x}=(x_1,\cdots,x_n)^T$ is a $n$-dimensional decision vector and $\mathbf{F}(\mathbf{x})$ is a $m$-dimensional objective vector. $\Omega\subseteq\mathbb{R}^n$ is the feasible region of the decision space, while $\mathbf{F}:\Omega\rightarrow\mathbb{R}^m$ is the corresponding attainable set in the objective space $\mathbb{R}^m$. Given two solutions $\mathbf{x}^1,\mathbf{x}^2 \in\Omega$, $\mathbf{x}^1$ is said to dominate $\mathbf{x}^2$, denoted by $\mathbf{x}^1\preceq\mathbf{x}^2$, if and only if $f_i(\mathbf{x}^1)\leq f_i(\mathbf{x}^2)$ for all $i\in\{1,\cdots,m\}$ and $\mathbf{F}(\mathbf{x}^1)\neq\mathbf{F}(\mathbf{x}^2)$. A solution $\mathbf{x}^*\in \Omega$ is said to be Pareto optimal if and only if no solution $\mathbf{x} \in \Omega$ dominates it. All Pareto optimal solutions constitute the Pareto-optimal set (PS) and the corresponding Pareto-optimal front (PF) is defined as $PF=\{\mathbf{F}(\mathbf{x})|\mathbf{x}\in PS\}$.

Evolutionary multi-objective optimization (EMO) algorithms, which are capable of approximating the PS and PF in a single run, have been widely accepted as a major approach for multi-objective optimization. Convergence and diversity are two cornerstones of multi-objective optimization: the former means the closeness to the PF while the latter indicates the spread and uniformity along the PF. Selection, which determines the survival of the fittest, plays a key role in balancing convergence and diversity. According to different selection mechanisms, the existing EMO algorithms can be roughly classified into three categories, i.e., Pareto-based methods~\cite{DebAPM02,SPEA2,KnowlesC00,LiDZZ16}, indicator-based methods~\cite{ZitzlerK04,BeumeNE07,BaderZ11} and decomposition-based methods~\cite{ZhangL07,LiFKZ14,LiZKLW14}.

This paper focuses on the decomposition-based methods, especially the multi-objective evolutionary algorithm based on decomposition (MOEA/D)~\cite{ZhangL07}. The original MOEA/D employs a steady-state selection mechanism, where the population is updated immediately after the generation of an offspring. In particular, this offspring is able to replace its neighboring parents when it has a better aggregation function value for the corresponding subproblem. To avoid a superior solution overwhelmingly occupying the whole population, \cite{LiZ09} suggested to restrict the maximum number of replacements taken by an offspring. More recently, \cite{LiZKLW14} developed a new perspective to understand the selection process of MOEA/D. Specifically, the selection process of MOEA/D is modeled as a one-one matching problem, where subproblems and solutions are treated as two sets of matching agents whose mutual preferences are defined as the convergence and diversity, respectively. Therefore, a stable matching between subproblems and solutions achieves an equilibrium between their mutual preferences, leading to a balance between convergence and diversity. However, as discussed in~\cite{LiKZD15} and~\cite{LiuGZ14}, partially due to the overrated convergence property, both the original MOEA/D and the stable matching-based selection mechanism fail to maintain the population diversity when solving problems with complicated properties, e.g., imbalanced problem~\cite{LiuGZ14,LiuCDG16} and many objectives. Bearing these considerations in mind, \cite{LiKZD15} modified the mutual preference definition and developed a straightforward but more effective selection mechanism based on the interrelationship between subproblems and solutions. Later on, \cite{WangZZGJ16} proposed an adaptive replacement strategy, which adjusts the replacement neighborhood size dynamically, to assign solutions to their most suitable subproblems. It is also interesting to note that some works took the advantages of the Pareto dominance- and decomposition-based selection mechanisms in a single paradigm~\cite{LiKWCR12,LiDZK15,CaiLFZ15,LiKD15, LiDZ15}.

To achieve a good balance between convergence and diversity, this paper suggests to introduce the concept of incomplete preference lists into the stable matching model. Specifically, borrowing the idea from the stable matching with incomplete preference lists~\cite{LiuMPS14}, we restrict the number of subproblems with which a solution is allowed to match. In this case, a solution can only be assigned to one of its favorite subproblems. However, due to the restriction on the preference list, the stable marriage model, which results in a one-one matching, may leave some subproblems unmatched. To remedy this situation, this paper implements two different versions of stable matching-based selection mechanisms with incomplete preference lists.
\begin{itemize}
	\item The first one achieves a two-level one-one matching. At the first level, we find the stable solutions for subproblems according to the incomplete preference lists. Afterwards, at the second level, the remaining unmatched subproblems are matched with suitable solutions according to the remaining preference information.
	\item The second one obtains a many-one matching. In such a way, the unmatched subproblems give the matching opportunities to other subproblems that have already matched with a solution but still have openings.
\end{itemize}
Note that the length of the incomplete preference list has a significant impact on the performance and is problem dependent~\cite{WuKZLWL15}. By analyzing the underlying mechanism of the proposed stable matching-based selection mechanisms in depth, we develop an adaptive mechanism to set the length of the incomplete list for each solution on the fly. Comprehensive experiments on 62 benchmark problems fully demonstrate the effectiveness and competitiveness of our proposed methods.

The rest of the paper is organized as follows. \pref{sec:preliminaries} introduces some preliminaries of this paper. Thereafter, the proposed algorithm is described step by step in~\pref{sec:proposal}. \pref{sec:setup} and \pref{sec:experiments} provide the experimental settings and the analysis of the empirical results. Finally, \pref{sec:conclusion} concludes this paper and provides some future directions.

% !Tex root = main.tex

\section{Preliminaries}
\label{sec:preliminaries}

In this section, we first introduce some background knowledge of MOEA/D and the stable matching-based selection. Then, our motivations are developed by analyzing their underlying mechanisms and drawbacks.

\subsection{MOEA/D}
\label{sec:moead}

As a representative of the decomposition-based algorithms, MOEA/D has become an increasingly popular choice for \textit{posterior} multi-objective optimization. Generally speaking, there are two basic components in MOEA/D: one is \textit{decomposition} and the other is \textit{collaboration}. The following paragraphs give some general descriptions of each component separately.

\subsubsection{Decomposition}
\label{sec:decomposition}

The basic idea of decomposition is transforming the original MOP into a single-objective optimization subproblem. There are many established decomposition methods developed for classic multi-objective optimization~\cite{nonlinear}, among which the most popular ones are weighted sum, Tchebycheff (TCH) and boundary intersection approaches. Without loss of generality, this paper considers the inverted TCH approach~\cite{LiZKLW14}, which is defined as follows:
\begin{equation}
\begin{array}{l}
\mathrm{minimize}\quad g^{tch}(\mathbf{x}|\mathbf{w},\mathbf{z}^{\ast})=\max\limits_{1\leq i\leq m}\{|f_i(\mathbf{x})-z_{i}^{\ast}|/{w_i}\}\\
\mathrm{subject\ to}\quad \mathbf{x}\in\Omega
\end{array},
\label{eq:TCH}
\end{equation}
where $\mathbf{w}=(w_1,\cdots,w_m)^T$ is a user specified weight vector, $w_i\geq 0$ for all $i\in\{1,\cdots,m\}$ and $\sum_{i=1}^{m}w_i=1$. In practice, $w_i$ is set to be a very small number, say $10^{-6}$, when $w_i=0$. $\mathbf{z}^{\ast}=(z^{\ast}_1,\cdots,z^{\ast}_m)^T$ is an Utopian objective vector where $z_i^{\ast}=\min\limits_{\mathbf{x}\in\Omega}f_i(x)$, $i\in\{1,\cdots,m\}$. Note that the search direction of the inverted TCH approach is $\mathbf{w}$, and the optimal solution of (\ref{eq:TCH}) is a Pareto-optimal solution of the MOP defined in (\ref{eq:MOP}) under some mild conditions. We can expect to obtain various Pareto-optimal solutions by using (\ref{eq:TCH}) with different weight vectors. In MOEA/D, a set of uniformly distributed weight vectors are sampled from a unit simplex.

\subsubsection{Collaboration}
\label{sec:collaboration}

As discussed in~\cite{ZhangL07}, the neighboring subproblems, associated with the geometrically close weight vectors, tend to share similar optima. In other words, the optimal solution of $g^{tch}(\cdot|\mathbf{w}^1,\mathbf{z}^{\ast})$ is close to that of $g^{tch}(\cdot|\mathbf{w}^2,\mathbf{z}^{\ast})$, given $\mathbf{w}^1$ and $\mathbf{w}^2$ are close to each other. In MOEA/D, each solution is associated with a subproblem. During the optimization process, the solutions cooperate with each other via a well-defined neighborhood structure and they solve the subproblems in a collaborative manner. In practice, the collaboration is implemented as a restriction on the mating and update procedures. More specifically, the mating parents are selected from neighboring subproblems and a newly generated offspring is only used to update its corresponding neighborhood. Furthermore, since different subproblems might have various difficulties, it is more reasonable to dynamically allocate the computational resources to different subproblems than treating all subproblems equally important. In~\cite{ZhangLL09}, a dynamic resource allocation scheme is developed to allocate more computational resources to those promising ones according to their online performance.

\subsection{Stable Matching-Based Selection}
\label{sec:stablematching}

Stable marriage problem (SMP) was originally introduced in~\cite{GaleS62} and its related work won the 2012 Nobel Prize in Economics. In a nutshell, the SMP is about how to establish a stable one-one matching between two sets of agents, say men and women, which have mutual preferences over each other. A stable matching should not contain a man and a woman who are not matched together but prefer each other to their assigned spouses.

In MOEA/D, subproblems and solutions can be treated as two sets of agents which have mutual preferences over each other. In particular, a subproblem prefers a solution that optimizes its underlying single-objective optimization problem as much as possible; while a solution prefers to have a well distribution in the objective space. The ultimate goal of selection is to select the best solution for each subproblem, and vice versa. In this case, we can treat the selection procedure as a one-one matching procedure between subproblems and solutions. To the best of our knowledge, MOEA/D-STM \cite{LiZKLW14} is the first one that has modeled the selection procedure of MOEA/D as an SMP, and encouraging results have been reported therein. The framework of the stable matching-based selection contains two basic components, i.e., \textit{preference settings} and \textit{matching model}. The following paragraphs briefly describe these two components.

\subsubsection{Preference Settings}

The preference of a subproblem $p$ on a solution $\mathbf{x}$ is defined as:
\begin{equation}
\Delta_P(p,\mathbf{x})=g^{tch}(\mathbf{x}|\mathbf{w},\mathbf{z}^{\ast}),
\label{eq:preference_p}
\end{equation}
where $\mathbf{w}$ is the weight vector of $p$. Consequently, $\Delta_P(p,\mathbf{x})$ measures the convergence of $\mathbf{x}$ with respect to $p$. The preference of a solution $\mathbf{x}$ to a subproblem $p$ is defined as:
\begin{equation}
\Delta_X(\mathbf{x},p)=\|\overline{F}(\mathbf{x})-\frac{\mathbf{w}^{\mathrm{T}}\cdot\overline{F}(\mathbf{x})}{\mathbf{w}^{\mathrm{T}}\cdot\mathbf{w}}\mathbf{w}\|,
\label{eq:preference_x}
\end{equation}
where $\overline{F}(\mathbf{x})$ is the normalized objective vector of $\mathbf{x}$ and $\|\cdot\|$ is the $\ell_2$-norm. Since the weight vectors are usually uniformly distributed, it is desirable that the optimal solution of each subproblem has the shortest perpendicular distance to its corresponding weight vector. For the sake of simplicity, $\Delta_X(\mathbf{x},p)$ can be used to measure the diversity of a solution~\cite{LiZKLW14}.

\subsubsection{Matching Model}

Based on the above preference settings, \cite{LiZKLW14} employed the classic deferred acceptance procedure (DAP) developed in~\cite{GaleS62} to find a stable matching between subproblems and solutions. The pseudo code of this stable matching-based selection mechanism is given in~\pref{alg:stm}. $\Psi_P$ and $\Psi_X$ are the preference matrices of subproblems and solutions, each row of which represents the preference list of a subproblem over all solutions, and vice versa. In particular, a preference list is built by sorting the preference values in an ascending order. $M$ indicates the set of all the constructed matching pairs. It is worth noting that the convergence and diversity have been aggregated into the preference settings, thus the stable matching between subproblems and solutions strikes the balance between convergence and diversity.

\begin{algorithm}[!t]
\caption{$\mathsf{STM}(P,S,\Psi_P,\Psi_X)$}
\label{alg:stm}
\KwIn{
\begin{itemize}
	\item subproblem set $P$ and solution set $S$
    \item sets of preference lists $\Psi_P$ and $\Psi_X$
\end{itemize}
}
\KwOut{stable matching set $M$}
$P_u\leftarrow P$, $M\leftarrow\emptyset$;\\
\While {$P_u\neq \emptyset$}{
    $p \leftarrow$ Randomly pick a subproblem from $P_u$;\\
    $\mathbf{x}\leftarrow$ First solution on $p$'s preference list;\\
    Remove $\mathbf{x}$ from $p$'s preference list;\\
    $M\leftarrow\mathsf{DAP}(p,\mathbf{x},P_u,M,\Psi_P,\Psi_X)$;\\
}
\Return $M$
\end{algorithm}

\begin{algorithm}[!t]
\caption{$\mathsf{DAP}(p,\mathbf{x},P_u,M,\Psi_P,\Psi_X)$}
\label{alg:DAP}
\KwIn{
\begin{itemize}
	\item current subproblem $p$ and solution $\mathbf{x}$
    \item unmatched subproblem set $P_u$
    \item current stable matching set $M$
    \item sets of preference lists $\Psi_P$ and $\Psi_X$
\end{itemize}
}
\KwOut{stable matching set $M$}
\uIf{$\mathbf{x}\notin M$}{
	$M \leftarrow M \cup (p,\mathbf{x})$\tcp*[r]{match $p$ and $\mathbf{x}$}
    $P_u\leftarrow P_u\setminus p$;\\
}
\Else{
    $p^\prime\leftarrow M(\mathbf{x})$\tcp*[r]{current partner of $\mathbf{x}$}
    \If{$\mathbf{x}$ prefers $p$ to $p^\prime$} {
    	$M \leftarrow M \cup (p,\mathbf{x}) \setminus (p^\prime,\mathbf{x})$;\\
        $P_u\leftarrow P_u \cup p^\prime \setminus p$;\\
    }
}
\Return $M$
\end{algorithm}

\subsection{Drawbacks of MOEA/D and MOEA/D-STM}
\label{sec:drawbacks}

In this subsection, we discuss some drawbacks of the selection mechanisms of MOEA/D and MOEA/D-STM.

\begin{figure}[!t]
	\centering
	\subfloat[MOEA/D]{\includegraphics[width=.3\linewidth]{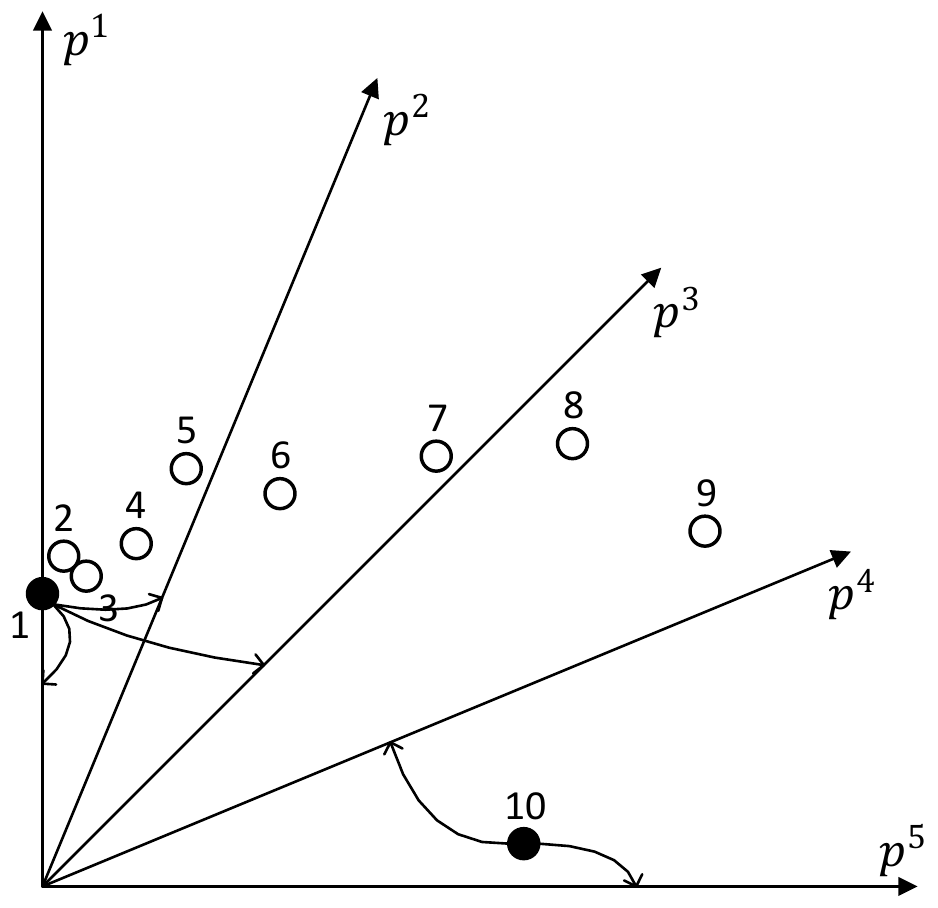}}
	\subfloat[MOEA/D-STM]{\includegraphics[width=.3\linewidth]{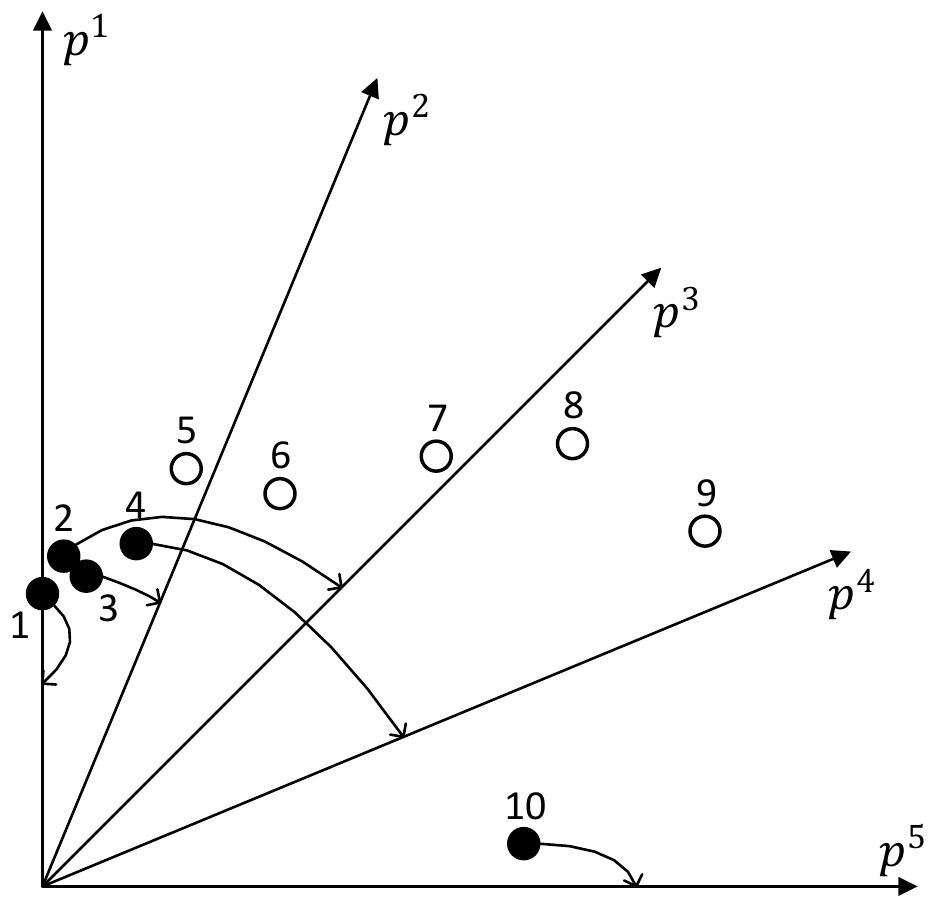}}
	\caption{Comparisons of MOEA/D and MOEA/D-STM.}
	\label{fig:selection}
\end{figure}

\subsubsection{MOEA/D}
\label{sec:drawbacks_MOEAD}

The update mechanism of the original MOEA/D is simple and efficient, yet greedy. In a nutshell, each subproblem simply selects its best solution according to the corresponding scalar optimization function value. As discussed in~\cite{ZhangLL09}, since different parts of the PF might have various difficulties, some subproblems might be easier than the others for finding the optimal solutions. During some intermediate stages of the optimization process, the currently elite solutions of some relatively easier subproblems might also be good candidates for the others. In this case, these elite solutions can easily take over all subproblems. In addition, it is highly likely that the offspring solutions generated from these elite solutions crowd into the neighboring areas of the corresponding subproblems. Therefore, this purely fitness-driven selection mechanism can be severely harmful for the population diversity and may lead to the failure of MOEA/D on some challenging problems~\cite{LiuGZ14}. Let us consider an example shown in~\pref{fig:selection}(a), where five out of ten solutions need to be selected for five subproblems. Since $\mathbf{x}^1$ is currently the best solution for $\{p^1,p^2,p^3\}$ and $\mathbf{x}^{10}$ is the current best candidate for $\{p^4,p^5\}$, these two elite solutions finally take over all five subproblems. Obviously, the population diversity of this selection result is not satisfied.

\subsubsection{MOEA/D-STM}
\label{sec:drawbacks_STM}

As discussed in~\cite{STMBook}, the DAP maximizes the satisfactions of the preferences of men and women in order to maintain the stable matching relationship. According to the preference settings for subproblems, solutions closer to the PF are always on the front of the subproblems' preference lists. In this case, the DAP might make some solutions match themselves with subprobolems lying on the rear of their preference lists. Even worse, as discussed in~\pref{sec:drawbacks_MOEAD}, these currently well converged solutions may crowd in a narrow area. This obviously goes against the population diversity. Let us consider the same example discussed in~\pref{fig:selection}(a). The preference matrices of subproblems and solutions are:

\begin{equation}
\Psi _P =
    \begin{array}{l}
        p^1:\\
        p^2:\\
        p^3:\\
        p^4:\\
        p^5:
    \end{array}
    \begin{matrix}
    [~1 & 2 & 3 & 4 & 5 & 6 & 7 & 8 & 10 & 9~] \\
    [~1 & 3 & 2 & 4 & 5 & 6 & 7 & 8 & 10 & 9~] \\
    [~1 & 3 & 2 & 4 & 6 & 5 & 7 & 10 & 8 & 9~] \\
    [~10 & 1 & 3 & 2 & 4 & 9 & 6 & 5 & 7 & 8~] \\
    [~10 & 1 & 3 & 2 & 4 & 9 & 6 & 5 & 7 & 8~]
    \end{matrix} \\
\label{eq:pps}
\end{equation}

\begin{equation}
\Psi _X =
    \begin{array}{l}
        \mathbf{x}^1:\\
        \mathbf{x}^2:\\
        \mathbf{x}^3:\\
        \mathbf{x}^4:\\
        \mathbf{x}^5:\\
        \mathbf{x}^6:\\
        \mathbf{x}^7:\\
        \mathbf{x}^8:\\
        \mathbf{x}^9:\\
        \mathbf{x}^{10}:\\
    \end{array}
    \begin{matrix}
    [~1 & 2 & 3 & 4 & 5~] \\
    [~1 & 2 & 3 & 4 & 5~] \\
    [~1 & 2 & 3 & 4 & 5~] \\
    [~2 & 1 & 3 & 4 & 5~] \\
    [~2 & 1 & 3 & 4 & 5~] \\
    [~2 & 3 & 1 & 4 & 5~] \\
    [~3 & 2 & 4 & 1 & 5~] \\
    [~3 & 4 & 2 & 5 & 1~] \\
    [~4 & 3 & 5 & 2 & 1~] \\
    [~5 & 4 & 3 & 2 & 1~]
    \end{matrix}~.
\label{eq:psp}
\end{equation}
From~\pref{eq:pps}, we can clearly see that $\mathbf{x}^1$ to $\mathbf{x}^4$ dominate the top positions of the preference lists of all subproblems. By using~\pref{alg:stm}, we have the selection/matching result shown in~\pref{fig:selection}(b), where $\mathbf{x}^1$ to $\mathbf{x}^4$ crowd in a narrow area between $p^1$ and $p^2$. This is obviously harmful for the population diversity as well.

From the above discussions, we find that the original selection mechanism of MOEA/D is a \textit{convergence first and diversity second} strategy~\cite{LiuGZ14}, which might give excessive priority to the convergence requirement. On the other hand, although the stable matching-based selection mechanism intends to achieve an equilibrium between convergence and diversity, the stable matching between subproblems and solutions may fail to keep the population diversity. This is because no restriction has been given to the subproblem with which a solution can match. In other words, a solution can match with an unfavorable subproblem in the resulting stable matching. To relieve this side effect, the next section suggests a strategy to take advantages of some partial information from the preference lists when finding the stable matching between subproblems and solutions.

% !Tex root = main.tex

\section{Adaptive Stable Matching-Based Selection with Incomplete Preference Lists}
\label{sec:proposal}

In the canonical SMP, each man/woman holds a complete and strictly ordered preference list over all agents from the other side. However, in practice, it may happen that a man/woman declares some unacceptable partners~\cite{STMBook}, and this results in an SMP with incomplete lists~\cite{IwamaMMM99}. By these means, a man/woman is only allowed to match with a matching agent that appears on his/her incomplete preference list. Due to the restriction from the incomplete preference lists, there is no guarantee that all agents can have a stable matching mate. A stable matching for an SMP with incomplete lists does not contain such a pair of man and woman: 1) they are acceptable to each other but not matched together; 2) they either do not match with anyone else or prefer each other to their current matching mates. To overcome the drawbacks discussed in~\pref{sec:drawbacks}, here we implement two versions of stable matching-based selection mechanisms with incomplete preference lists: one achieves a two-level one-one matching while the other obtains a many-one matching.

\subsection{Two-Level One-One Stable Matching-Based Selection}
\label{sec:twolevelstm}

\begin{algorithm}[!t]
	\caption{$\mathsf{STMIC}(P,S,\Psi _P,\Psi _X,R)$}
	\label{alg:stmic}
	\KwIn{
		\begin{itemize}
			\item subproblem set $P$, solution set $S$
			\item preference matrices $\Psi_P$ and $\Psi_X$
			\item length of solution's preference list set $R$
		\end{itemize}
	}
	\KwOut{stable matching set $M$}
	$P_u\leftarrow P$, $M\leftarrow\emptyset$;\\
	\For{$i \leftarrow 1$ \emph{to} $|S|$}{
		Keep the first $r^i$ subproblems on $\mathbf{x}^i$'s complete preference list and remove the remainders;\\
	}
	\While {$P_u\neq \emptyset$}{
		$p \leftarrow$ Randomly select a subproblem from $P_u$;\\
		\uIf{$p$'s preference list $\neq \emptyset$} {
			$\mathbf{x} \leftarrow$ First solution on $p$'s preference list;\\
			Remove $\mathbf{x}$ from $p$'s preference list;\\
			\If{$p$ is on $\mathbf{x}$'s preference list} {
				$M\leftarrow\mathsf{DAP}(p,\mathbf{x},P_u,M,\Psi_P,\Psi_X)$;\\
			}
		}
		\Else{
			$P_u\leftarrow P_u\setminus p$;\\
		}
	}
	\textbf{Return} $M$;\\
\end{algorithm}

\begin{algorithm}[!t]
	\caption{$\mathsf{SelectionOOSTM2L}(P,S)$}
	\label{alg:stm2l}
	\KwIn{subproblem set $P$ and solution set $S$
	}
	\KwOut{solution set $S$}
	\tcc{First-level stable matching}
	Compute $\Psi _p$ and $\Psi _\mathbf{x}$ for $P$ and $S$;\\
	$R\leftarrow$ Set the length of each solution's preference list;\\
	$M\leftarrow$ $\mathsf{STMIC}(P,S,\Psi _P,\Psi _X,R)$;\\
	\tcc{Second-level stable matching}
	$(P_m,S_m)\leftarrow M$;\\
	$P_u\leftarrow P\setminus P_m$;\\
	$S_u\leftarrow S\setminus S_m$;\\
	Compute $\Psi _P^\prime$ and $\Psi _X^\prime$ for $P_u$ and $S_u$;\\
	$M^\prime \leftarrow$ $\mathsf{STM}(P_u,S_u,\Psi _P^\prime,\Psi _X^\prime)$;\\
	\tcc{Combine the stable matching pairs}
	$M\leftarrow M\cup M^\prime$;\\
	\textbf{Return} $M$;\\
\end{algorithm}

In the first level, let us assume that there are $N$ subproblems and $Q$ solutions, where $N<Q$. After obtaining the complete preference lists of all subproblems and solutions (line 1 and line 2 of~\pref{alg:stm2l}), we only keep the first $r^i$, where $i\in\{1,\cdots,Q\}$ and $0<r^i\leq N$, subproblems on the preference list of each solution $\mathbf{x}^i$, while the remaining ones are not considered any longer (line 2 and line 3 of~\pref{alg:stmic}). In this case, each solution is only allowed to match with its first several favorite subproblems which are close to itself according to~\pref{eq:preference_x}. In contrast, the preference lists of subproblems are kept unchanged. Given the incomplete preference information, we employ the DAP to find a stable matching between subproblems and solutions (line 4 to line 12 of~\pref{alg:stmic}). By these means, we can expect that the population diversity is strengthened during the first-level stable matching. This is because a solution is not allowed to match with an unfavorable subproblem which lies out of its incomplete preference list. The pseudo code of the stable matching with incomplete lists is given in~\pref{alg:stmic}.

During the first-level stable matching, not all subproblems are assigned with a stable solution due to the incomplete preference information. To remedy this issue, the second-level stable matching with complete preference lists is developed to find a stable solution for each unmatched subproblem. At first, we compute the preference matrices of the unmatched subproblems and solutions (line 7 of~\pref{alg:stm2l}). Afterwards, we employ \pref{alg:stm} to find a stable matching between them (line 8 of~\pref{alg:stm2l}). In the end, the matching pairs of both levels of stable matching are gathered together to form the final selection results (line 9 of~\pref{alg:stm2l}). The pseudo code of the two-level stable matching-based selection mechanism is given in~\pref{alg:stm2l}.

\subsection{Many-One Stable Matching-Based Selection}
\label{sec:manyonestm}

Many-one stable matching problem is an extension of the standard SMP, where a matching agent from one side is allowed to have more than one matching mates from the other side. For example, in the college admission problem (CAP)~\cite{GaleS62}, the colleges and applicants are two sets of matching agents. Each college has a preference list over all applicants and vice versa. Different from the SMP, each applicant is only allowed to enter one college, whereas each college has a positive integer quota being the maximum number of applicants that it can admit.

As the other implementation of the stable matching-based selection with incomplete preference lists, here we model the selection process of MOEA/D as a CAP with a common quota~\cite{BiroFIM10}. More specifically, subproblems and solutions are treated as colleges and applicants respectively. A solution is only allowed to match with one subproblem while a subproblem is able to match with more than one solution. In particular, we do not limit the separate quota for every subproblem but assign a common quota for all subproblems, which equals the number of subproblems (i.e., $N$). In other words, $N$ subproblems can at most match with $N$ solutions in this many-one matching. Note that a matching is stable if there does not exist any pair of subproblem $p$ and solution $\mathbf{x}$ where:
\begin{itemize}
	\item $p$ and $\mathbf{x}$ are acceptable to each other but not matched together;
	\item $\mathbf{x}$ is unmatched or prefers $p$ to its assigned subproblem;
	\item the common quota is not met or $p$ prefers $\mathbf{x}$ to at least one of its assigned solutions.
\end{itemize} 

\begin{algorithm}[!t]
	\caption{$\mathsf{SelectionMOSTM}(P,S)$}
	\label{alg:mostmic}
	\KwIn{subproblem set $P$ and solution set $S$}
	\KwOut{stable matching set $M$}
	Compute $\Psi _P$ and $\Psi _X$ for $P$ and $S$;\\
	$R\leftarrow$ Set the length of each solution's preference list;\\
	$S_u\leftarrow S$, $M\leftarrow\emptyset$;\\
	\For{$i \leftarrow 1$ \emph{to} $Q$}{
		Keep the first $r^i$ subproblems on $\mathbf{x}^i$'s complete preference list and remove the remainders;\\
	}
	\While {$S_u\neq \emptyset$}{
		$\mathbf{x}\leftarrow$ Randomly select a solution from $S_u$;\\
		\uIf{$\mathbf{x}$'s preference list $\neq \emptyset$} {
			$p\leftarrow$ First subproblem on $\mathbf{x}$'s preference list;\\
			Remove $p$ from $\mathbf{x}$'s preference list;\\
			$M\leftarrow M\cup(p,\mathbf{x})$;\\
			\If{$|M|>N$}{
				$\overline{P}\leftarrow\mathop{\arg\max}\limits_{p\in P}|M(p)|$\tcp*[r]{$|M(p)|$ is the cardinality of $M(p)$}
				$\overline{P}\leftarrow\mathop{\arg\max}\limits_{p\in \overline{P}}\{\max\limits_{\mathbf{x}\in M(p)}
				{rank(p,\mathbf{x})}\}$\tcp*[r]{$rank(p,\mathbf{x})$ is the rank of $\mathbf{x}$ on $p$'s preference list}
				$p^\prime\leftarrow$ Randomly select a subproblem from $\overline{P}$;\\
				$\mathbf{x}^\prime\leftarrow \mathop{\arg\max}\limits_{\mathbf{x}\in M(p^\prime)}rank(p^\prime,\mathbf{x})$;\\
				$M\leftarrow M\setminus(p^\prime,\mathbf{x}^\prime)$;\\
				$S_u\leftarrow S_u\cup\mathbf{x}^\prime$;\\
			}
		}
		\Else{
			$S_u\leftarrow S_u\setminus\mathbf{x}$;\\
		}
	}
	\Return $M$;\\
\end{algorithm}

The pseudo code of the many-one stable matching-based selection mechanism is given in~\pref{alg:mostmic}. The initialization process (line 1 to line 5 of~\pref{alg:mostmic}) is the same as the one-one stable matching discussed in~\pref{sec:twolevelstm}. During the main while-loop, an unmatched solution $\mathbf{x}\in S_u$ at first matches with its current favorite subproblem $p$ according to its preference list (line 7 to line 11 of~\pref{alg:mostmic}). If the number of current matching pairs $|M|$ is larger than $N$, we find a substitute subproblem $p^\prime$ and adjust its matching pairs by releasing the matching relationship with its least preferred solution $\mathbf{x}^\prime$ (line 12 to line 18 of~\pref{alg:mostmic}). In particular, $p^\prime$ is selected according to the following criteria:
\begin{itemize}
\item At first, we choose the subproblems that have the largest number of matched solutions to form $\overline{P}$ (line 13 of~\pref{alg:mostmic}). Its underlying motivation is to reduce the chance for overly exploiting a particular subproblem.
\item If the cardinality of $\overline{P}$ is greater than one, we need to further process $\overline{P}$. Specifically, we investigate the ranks of the solutions matched with subproblems in $\overline{P}$. The subproblems, whose least preferred solution holds the worst rank on that subproblem's preference list, are used to reconstruct $\overline{P}$ (line 14 of~\pref{alg:mostmic}).
\item In the end, $p^\prime$ is randomly chosen from $\overline{P}$ (line 14 of~\pref{alg:mostmic}).
\end{itemize}
Note that we add $\mathbf{x}^\prime$ back into $S_u$ after releasing its matching relationship with $p^\prime$ (line 18 of~\pref{alg:mostmic}). The matching process terminates when $S_u$ becomes empty.

\subsection{Impacts of the Length of the Incomplete Preference List}
\label{sec:preference_length}

\begin{figure*}[!t]
	\centering
	\subfloat[STM2L ($r=4$)]{\includegraphics[width=.25\linewidth]{Figs/Drawing_stm.pdf}}
	\subfloat[STM2L ($r=3$)]{\includegraphics[width=.25\linewidth]{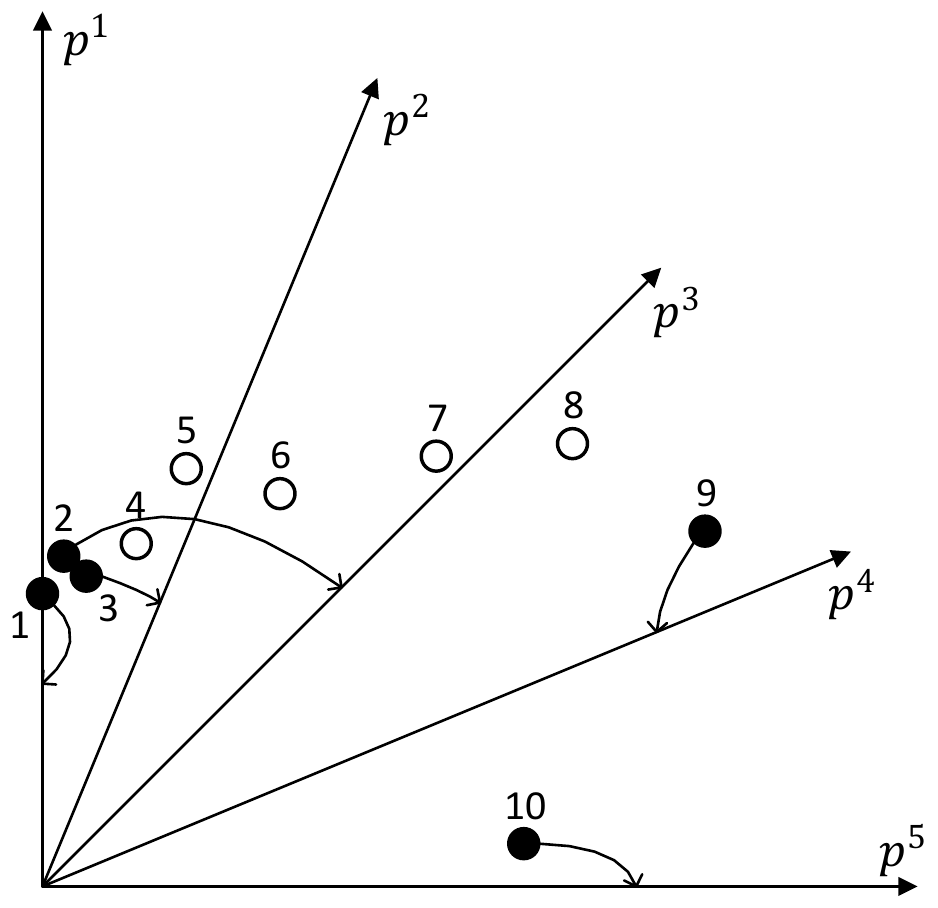}}
	\subfloat[STM2L ($r=2$)]{\includegraphics[width=.25\linewidth]{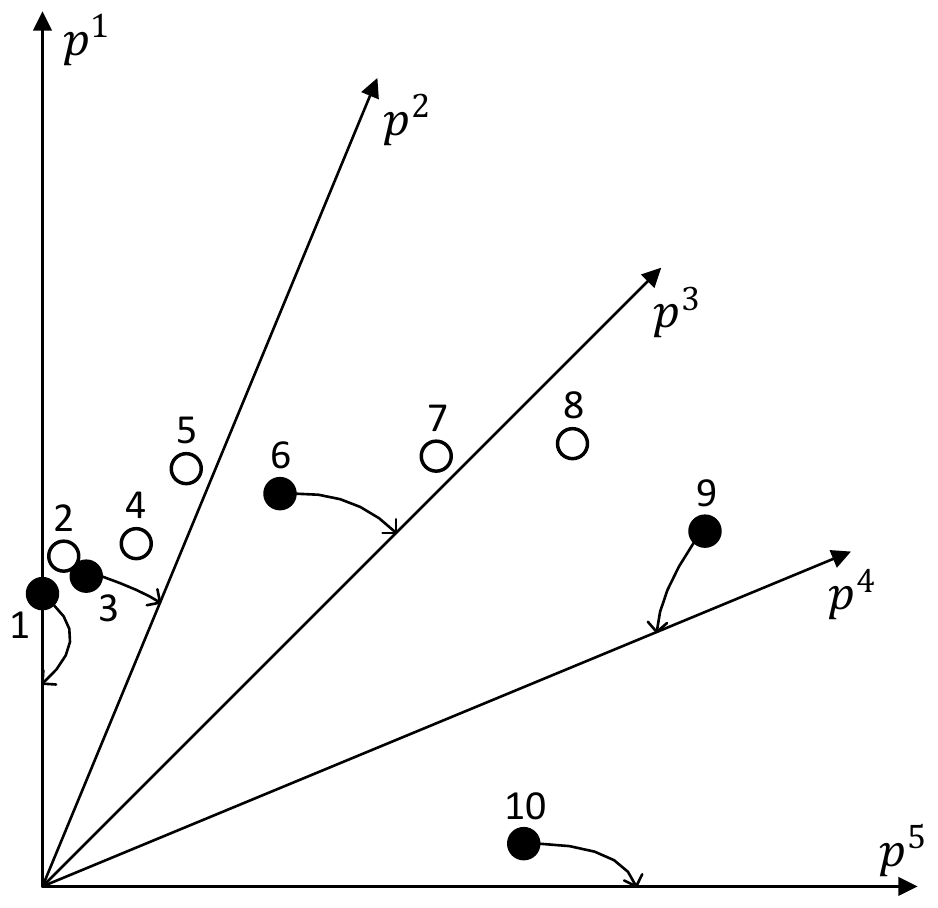}}
	\subfloat[STM2L ($r=1$)]{\includegraphics[width=.25\linewidth]{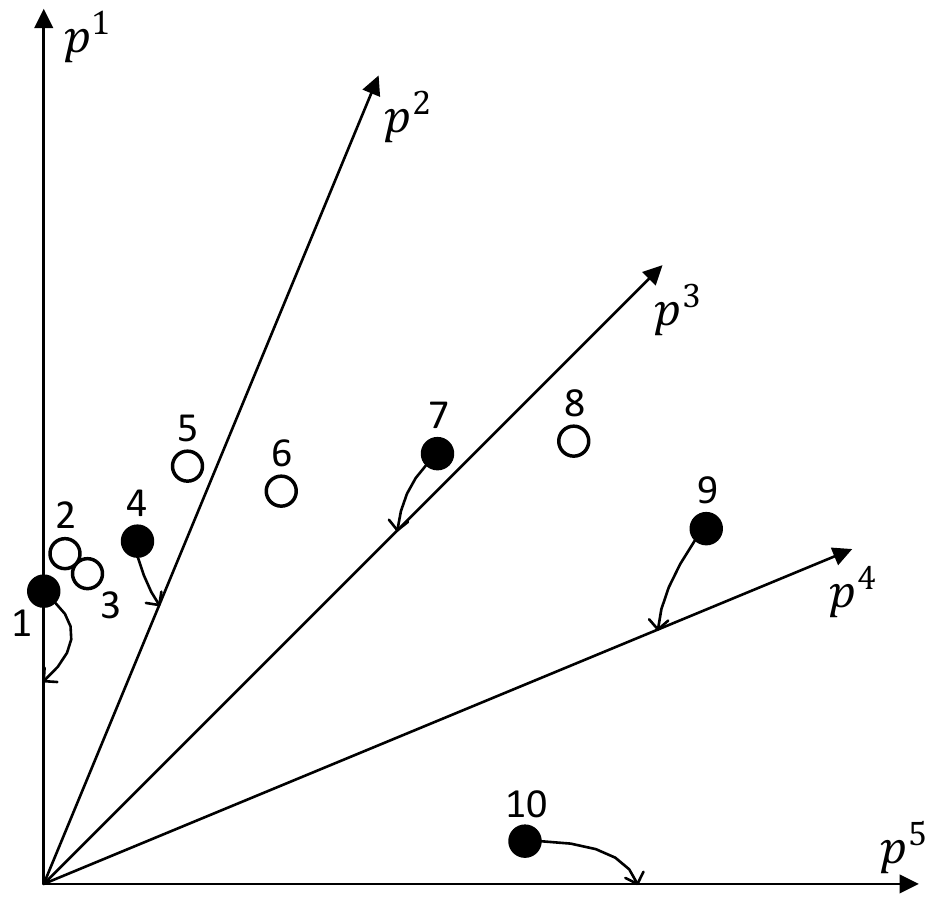}}
	\caption{Comparisons of the two-level stable matching-based selection with incomplete lists using different $r$ settings.}
	\label{fig:stm2l-eg}
\end{figure*}

As discussed in the previous subsections, we expect to improve the population diversity by restricting the length of the preference list of each solution. A natural question is whether this length affects the behavior of our proposed stable matching-based selection mechanisms? Let us consider the example discussed in~\pref{fig:selection} again. For the sake of discussion, here we set the length of the incomplete preference list of each solution as a constant (denoted by $r$). By using different settings of $r$, \pref{fig:stm2l-eg} shows the selection results of the two-level stable matching-based selection mechanism. From this figure, we find that the diversity of the selected solutions increases with the decrease of $r$; on the other hand, the improvement of the diversity is at the expense of the convergence. It is interesting to note that the two-level stable matching-based selection mechanism totally degenerates into the original stable matching-based selection mechanism shown in~\pref{fig:selection}(a) when using $r=4$. In a word, $r$ controls the trade-off between convergence and diversity in the stable matching-based selection with incomplete preference lists. In the next subsection, we develop an adaptive mechanism to control the length of each solution's preference list on the fly.

\subsection{Adaptive Mechanism}
\label{sec:adaptive}

To better understand the proposed adaptive mechanism, here we introduce the concept of \textit{local competitiveness}. At first, all solutions are associated with their closest subproblems having the shortest perpendicular distance between the objective vector of the solution and the weight vector of the subproblem. Afterwards, for each subproblem having more than one associated solutions, we choose the one, which has the best aggregation function value, as its representative solution. A solution is defined as a \textit{locally competitive} solution in case it dominates at least one representative solution of its $\ell\geq 1$ nearest subproblems; otherwise, it is defined as a \textit{locally noncompetitive} solution. In view of the population dynamics of the evolutionary process, we develop an adaptive mechanism to set the length of the incomplete preference list of a solution according to its local competitiveness (\pref{alg:r} gives its pseudo code). Briefly speaking, this length is set as the maximum $\ell$ that keeps the corresponding solution locally noncompetitive.

More specifically, given $N$ subproblems and $Q$ solutions, each solution is associated with its closest subproblem as shown in line 1 and line 2 of~\pref{alg:r}. In particular, $\Phi[i]$ represents the index of the subproblem with which a solution $\mathbf{x}^i$ is associated, $i\in\{1,\cdots,Q\}$. In line 4 of~\pref{alg:r}, we collect the associated solutions of each subproblem $p^j$, $j\in\{1,\cdots,N\}$, to form a temporary set $\chi$. Then, line 5 to line 8 of~\pref{alg:r} determine the representative solution of each subproblem $p^j$, where $\varphi[j]$ represents the index of its representative solution. Afterwards, for each solution $\mathbf{x}^i$, line 10 to line 16 of~\pref{alg:r} gradually increase $\ell$ until $\mathbf{x}^i$ becomes locally competitive, and this final $\ell$ is used as the length of $\mathbf{x}^i$'s incomplete preference list. Note that since each solution locates within the subspace between $m$ closest neighboring weight vectors in $m$-dimensional objective space, it can be associated with any of these $m$ subproblems in principle. Moreover, to avoid unnecessary comparisons, it is desirable to keep the solution's incomplete preference list within a reasonably small length. All in all, the length of $\mathbf{x}^i$'s incomplete preference list is adaptively tuned between $m$ and $\ell_{max}$. In particular, $\ell_{max}$ is set as the neighborhood size $T$ used in MOEA/D, where the mating parents are selected from.

\begin{algorithm}[!t]
	\caption{$\mathsf{AdaptiveSetR(P,S,\Psi_X)}$}
	\label{alg:r}
	\KwIn{
		\begin{itemize}
			\item subproblem set $P$ and solution set $S$
			\item solution preference matrix $\Psi_X$
		\end{itemize}
	}
	\KwOut{length of solution's preference list set $R$}
	
	\For{$i\leftarrow 1$ \KwTo$Q$} {
		$\Phi[i] \leftarrow \Psi _X[i][1]$;\\
	}
	\For{$j\leftarrow 1$ \KwTo $N$} {
		$\chi\leftarrow \{i|\Phi[i]=j,~i\in {1,2,...,Q}\}$;\\
		\uIf{$\chi=\emptyset$} {
			$\varphi[j]\leftarrow -1$;\\
		}
		\Else {
			$\varphi[j]\leftarrow\mathop{\arg\min}\limits_{i\in\chi}g^{tch}(\mathbf{x}^i|\lambda^j,\mathbf{z}^{\ast})$;\\
		}
	}
	\For{$i\leftarrow 1$ \KwTo $Q$} {
		$r^i\leftarrow m$;\\
		\For {$\ell\leftarrow m+1$ \KwTo $\ell_{max}$}{
			$t \leftarrow \varphi[\Psi _X[i][\ell]]$;\\
			\If{$t\neq -1$}{
				\If{$\mathbf{x}^i \prec \mathbf{x}^{t}$}{
					break;\\
				}
			}
			$r^i\leftarrow\ell$;\\
		}
	}
	\Return $R$;\\
\end{algorithm}

Let us use the example shown in~\pref{fig:selection}(b) to explain the underlying principle of our proposed adaptive mechanism. In this example, solutions $\mathbf{x}^2$ and $\mathbf{x}^3$ become locally competitive when $\ell >1$; while solutions $\mathbf{x}^7$ and $\mathbf{x}^9$ are locally noncompetitive for all $\ell$ settings. It is worth noting that neither $\mathbf{x}^2$ nor $\mathbf{x}^3$ is the representative solution of any subproblem; in the meanwhile, they are crowded in a narrow area. Since these locally competitive solutions have better ranks in the preference lists than those less competitive ones, the original stable matching-based selection tends to give them higher priorities to form the matching pairs. However, this selection result is obviously harmful for the population diversity. In addition, we also notice that $\mathbf{x}^7$ and $\mathbf{x}^9$ are the representative solutions of $p^3$ and $p^4$, thus they should contain some relevant information for optimizing these subproblems. In contrast, although $\mathbf{x}^2$ and $\mathbf{x}^3$ have better aggregation function values, they are far away from $p^3$ and $p^4$ and should be less relevant to them. To resolve these issues, our proposed adaptive mechanism adaptively restricts the length of the preference list of each solution $\mathbf{x}$ by removing subproblems whose representative solution is dominated by $\mathbf{x}$. By these means, we can make sure that each solution does not consider a subproblem which prefers this solution to its own representative solution. Thus each subproblem is prevented from matching with a less relevant solution. Note that this adaptive mechanism can be readily plugged into both of our proposed two versions of stable matching-based selection mechanisms by using~\pref{alg:r} to replace line 2 of \pref{alg:stm2l} and \pref{alg:mostmic}, respectively. The adaptive two-level one-one stable matching-based selection mechanism and the adaptive many-one stable matching-based selection mechanism are denoted by AOOSTM and AMOSTM for short.

\subsection{Time Complexity of AOOSTM and AMOSTM}
\label{sec:complexity}

In this subsection, we analyze the complexity of AOOSTM and AMOSTM. For both selection mechanisms, the calculation of $\Delta_P(p,\mathbf{x})$ and $\Delta_X(\mathbf{x},p)$ cost $\mathcal{O}(NQlogQ)$ computations \cite{LiZKLW14}. In \pref{alg:r}, the association operation between subproblems and solutions costs $\mathcal{O}(Q)$ calculations (line 1 to line 2). As for line 3 to line 8 of \pref{alg:r}, the identification of the representative solution for each subproblem requires $\mathcal{O}(mNQ)$ computations. Thereafter, the computation of $R$ in line 9 to line 16 of \pref{alg:r} costs $\mathcal{O}(mQ(\ell_{max}-m))$ computations in the worst case. Considering the two-level one-one stable matching in \pref{alg:stm2l}, the complexity of the one-one stable matching with the incomplete lists in line 3 is $\mathcal{O}(N\ell_{max})$, which is simpler than the original stable matching with complete preference lists \cite{LiZKLW14}. Next, the complexity of line 4 to line 6 of \pref{alg:stm2l} is $\mathcal{O}(N+Q)$. During the second-level stable matching (line 7 to line 8 of \pref{alg:stm2l}), same complexity analysis can be done for the remaining subproblems and solutions. Overall, the total complexity of AOOSTM is $\mathcal{O}(\max(NQlogQ,mQ(\ell_{max}-m)))$. When it comes to AMOSTM, since \pref{alg:mostmic} is solution-oriented, the computational complexity of line 3 to line 19 is $\mathcal{O}(Q\ell_{max})$. The total complexity of AMOSTM is still $\mathcal{O}(\max(NQlogQ,mQ(\ell_{max}-m)))$.

\subsection{Incorporation with MOEA/D}
\label{sec:moead-astm}

Similar to \cite{LiZKLW14}, we choose the MOEA/D-DRA~\cite{ZhangLL09} as the base framework and replace the update mechanism by the AOOSTM and AMOSTM selection mechanisms developed in~\pref{sec:adaptive}. The resulted algorithms are denoted by MOEA/D-AOOSTM and MOEA/D-AMOSTM, of which the pseudo code is given in~\pref{alg:moead-astm}. Note that the normalization scheme proposed in~\cite{DebJ14} is adopted to handle MOPs with different scales of objectives. In the following paragraphs, some important components of MOEA/D-AOOSTM/AMOSTM are further illustrated.

\begin{algorithm}[!t]
	\caption{$\mathsf{MOEA/D}$--$\mathsf{AOOSTM/AMOSTM}$}
	\label{alg:moead-astm}
	
	\KwIn{algorithm parameters}
	\KwOut{final population $S$}
	Initialize the population $S$, a set of weight vectors $W$ and their neighborhood structure $B$;\\
	$M\leftarrow$ Random one-one matching between $P$ and $S$;\\
	$neval\leftarrow 0$, $iteration\leftarrow 0$;\\
	\While{Stopping criterion is not satisfied}{
		Select the current active subproblems to form $I$;\\
		\For{each $i \in I$}{
			\uIf{$uniform(0,1)<\delta$ {\&\&} $|E|>=T$}{
				$E\leftarrow \{M(p)|p\in B(i)\}$;\\
			}
			\Else{
				$E\leftarrow S$; \\
			}
			Randomly select mating solutions from $E$ and generate an offspring $\mathbf{\overline{x}}$, $S\leftarrow S\cup\mathbf{\overline{x}}$;\\
			Evaluate $\mathbf{F}(\mathbf{\overline{x}})$, $neval$++;\\
		}
		$M\leftarrow\mathsf{SelectionOOSTM2L/MOSTM}(P,S)$;\\
		$S\leftarrow\{M(p)|p\in P\}$;\\
		$iteration$++;\\
		\If{$mod(iteration,30)=0$}{
			Update the utility of each subproblem;\\
		}
	}
	\Return $S$;
\end{algorithm}

\subsubsection{Initialization}
\label{sec:initialization}

Without any prior knowledge of the landscape, the initial population $S=\{\mathbf{x}^1,\cdots,\mathbf{x}^N\}$ is randomly sampled from $\Omega$. Same as the original MOEA/D, we use the classic method suggested in~\cite{NBI} to generate a set of uniformly distributed weight vectors $W=\{\mathbf{w}^1,\cdots,\mathbf{w}^N\}$ on a unit simplex. In addition, for each weight vector $\mathbf{w}^i$, $i\in\{1,\cdots,N\}$, we assign its $T$, $1\leq T\leq N$, closest weight vectors as its neighbors.

\subsubsection{Reproduction}
\label{sec:reproduction}

According to the underlying test problem, here we employ the widely used differential evolution (DE)~\cite{DEsurvey}, simulated binary crossover (SBX)~\cite{agrawal1994simulated} and polynomial mutation~\cite{PolyMutation} for offspring generation.

\subsubsection{Utility of Subproblem~\cite{ZhangLL09}}

The utility of subproblem $p^i$, denoted by $\pi^{i}$, $i\in\{1,\cdots,N\}$, measures the improvement rate of $p^i$. We make some modifications on $\pi^i$ to fit our proposed MOEA/D-AOOSTM/AMOSTM:
%how much improvement has been achieved by its current solution $\mathbf{x}^{new}$ in reducing the inverted TCH value of $p^i$.
\begin{equation}
\pi^{i}=
\begin{cases}
1 & \text{if}\ \Delta^{i}>0.001\\
0.95\times\pi^{i} & \text{if}\ \Delta^{i}<0\\
(0.95+0.05\times\frac{\Delta^{i}}{0.001})\times\pi^{i} & \text{otherwise}
\end{cases}
\end{equation}
where $\Delta^{i}$ represents the relative decrease of the scalar objective value of $p^i$ and is evaluated as:
\begin{equation}
%\Delta^{i}=\frac{g(\mathbf{x}^{i,old}|\mathbf{w}^i,\mathbf{z}^{\ast})-g(\mathbf{x}^{i,new}|\mathbf{w}^i,\mathbf{z}^{\ast})}{g(\mathbf{x}^{i,old}|\mathbf{w}^i,\mathbf{z}^{\ast})}\\
\Delta^{i}=
\begin{cases}
\frac{g^{tch}(\mathbf{x}^{i,old}|\mathbf{w}^i,\mathbf{z}^{\ast})-g^{tch}(\mathbf{x}^{i,best}|\mathbf{w}^i,\mathbf{z}^{\ast})}{g^{tch}(\mathbf{x}^{i,old}|\mathbf{w}^i,\mathbf{z}^{\ast})} & \text{if}\ M(p^i)\neq\emptyset\\
0 & \text{otherwise}
\end{cases}
\end{equation}
where $\mathbf{x}^{i,best}=\mathop{\arg\min}\limits_{\mathbf{x}\in M(p^i)}g^{tch}(\mathbf{x}|\mathbf{w}^i,\mathbf{z}^{\ast})$ is the best solution matched with $p^i$ in the current generation and $\mathbf{x}^{i,old}$ is the previously saved value of $\mathbf{x}^{i,best}$.

% !Tex root = main.tex

\section{Experimental Settings}
\label{sec:setup}

This section presents the general setup of our empirical studies, including the benchmark problems, algorithms in comparisons, parameter settings and performance metrics.

\subsection{Benchmark Problems}
\label{sec:benchmark}

From three popular benchmark suites, i.e., MOP~\cite{LiuGZ14}, UF~\cite{zhang2008multiobjective} and WFG~\cite{HubandHBW06}, 62 problem instances in total, are chosen as the benchmark set in our empirical studies. These problem instances have various characteristics, e.g., non-convexity, deceptive, multi-modality. According to the recommendations in the original references, the number of decision variables is set as: $n=10$ for the MOP instances and $n=30$ for the UF instances. As the WFG instances are scalable to any number of objectives, here we consider $m\in\{2,3,5,8,10\}$. In particular, when $m=2$, $n=k+l$~\cite{LiKZD15}, where the position-related variable $k=2$ and the distance-related variable $l=4$; while for $m\geq 3$, we use the recommended settings in~\cite{LiDZK15} and~\cite{DebJ14}, i.e., $k=2\times (m-1)$ and $l=20$.

\subsection{Algorithms in Comparisons}
\label{sec:emos}

Nine state-of-the-art EMO algorithms, i.e., MOEA/D-STM, MOEA/D-IR~\cite{LiKZD15}, gMOEA/D-AGR~\cite{WangZZGJ16}, MOEA/D-M2M~\cite{LiuGZ14}, MOEA/D-DRA, HypE~\cite{BaderZ11}, NSGA-III~\cite{DebJ14}, PICEA-g~\cite{WangPF13} and MOEA/DD~\cite{LiDZK15}, are considered in our empirical studies. In particular, the first seven algorithms are used for comparative studies on problems with complicated PSs; while the latter five are chosen to investigate the scalability on problems with more than three objectives. The characteristics of these algorithms are briefly described in the supplementary file of this paper\footnote{https://coda-group.github.io/publications/suppASTM.pdf}.

\subsection{Parameter Settings}
\label{sec:parameters}

\begin{table}[!t]
	\centering
	\caption{Settings of Population Size.}
	\label{tab:popsize}
	\begin{tabular}{c|c|c}
		\toprule
		Benchmark Problem & $m$ & Population Size	\\\midrule
		UF1 to UF7 & 2 & 600	\\\midrule
		UF8 to UF10 & 3 & 1,000	\\\midrule
		MOP1 to MOP5 & 2 & 100	\\\midrule
		MOP6 to MOP7 & 3 & 300	\\\midrule
		WFG1 to WFG9 & 2 & 250	\\\midrule
		WFG1 to WFG9 & 3 & 91	\\\midrule
		WFG1 to WFG9 & 5 & 210	\\\midrule
		WFG1 to WFG9 & 8 & 156	\\\midrule
		WFG1 to WFG9 & 10 & 275	\\\bottomrule
	\end{tabular}
\end{table}

Referring to \cite{LiZKLW14,LiKZD15} and \cite{LiDZK15}, the settings of the population size $N$ for different benchmark problems are shown in \pref{tab:popsize}. The stopping condition of each algorithm is the predefined number of function evaluations. In particular, it is set to $300,000$ for the UF and MOP instances~\cite{LiZKLW14}, and $25,000$ for the bi-objective WFG instances~\cite{LiKZD15}. As for the many-objective WFG instances, where $m\in\{3,5,8,10\}$, the number of function evaluations is set as $91\times N$, $210\times N$, $156\times N$ and $275\times N$, respectively~\cite{LiDZK15}. The parameters of our proposed MOEA/D-AOOSTM and MOEA/D-AMOSTM are set as follows:
\begin{itemize}
	\item\textit{Reproduction operators}: As for problems with complicated properties, we use the DE operator and polynomial mutation for offspring generation. As recommended in~\cite{LiKZD15}, we set $CR=1.0$ and $F=0.5$ for the UF and MOP instances; while $CR=F=0.5$ for bi-objective WFG instances. The mutation probability $p_m$ is set to be $\frac{1}{n}$ and its distribution index $\eta_m$ equals $20$. For problems with more than three objectives, we use the SBX operator to replace the DE operator, where the crossover probability $p_c=1$ and its distribution index $\eta_c=30$~\cite{DebJ14}. All other MOEA/D variants in our experimental studies share the same settings for reproduction operators.
	\item\textit{Neighborhood size}: $T=20$~\cite{LiZKLW14,LiKZD15}.
	\item\textit{Probability to select $B$ in the neighborhood}: $\delta=0.9$~\cite{ZhangLL09}.
%	\item Upper limit of $\ell$: $\ell_{max}=T=20$.
\end{itemize}

\subsection{Performance Metrics}
\label{sec:metrics}

To assess the performance of different algorithms, we choose the following two widely used performance metrics:
\begin{enumerate}
	\item\textit{Inverted Generational Distance} (IGD)~\cite{BosmanT03}: Given $P^*$ as a set of points uniformly sampled along the PF and $P$ as the set of solutions obtained from an EMO algorithm. The IGD value of $P$ is calculated as:
	\begin{equation}
	IGD(P,P^*)=\frac{\sum_{\mathbf{z}\in P^*}dist(\mathbf{z},P)}{|P^*|},
	\end{equation}
	where $dist(\mathbf{z},P)$ is the Euclidean distance of $\mathbf{z}$ to its nearest point in $P$.
	\item\textit{Hypervolume} (HV)~\cite{ZitzlerT99}: Let $\mathbf{z}^r=(z^r_1,\cdots,z^r_m)^T$ be a point dominated by all the Pareto optimal objective vectors. The HV of $P$ is defined as the volume of the objective space dominated by the solutions in $P$ and bounded by $\mathbf{z}^r$:
	\begin{equation}
	HV(P)=\textsf{VOL}(\bigcup \limits_{\mathbf{z}\in P} [z_1,z^r_1]\times\cdots\times [z_m,z^r_m]),
	\end{equation}
	where \textsf{VOL} indicates the Lebesgue measure.
\end{enumerate}

Since the objective functions of WFG instances are in different scales, we normalize their PFs and the obtained solutions in the range of $[0,1]$ before calculating the performance metrics. In this case, we constantly set $\mathbf{z}^r=(1.2,\cdots,1.2)^T$ in the HV calculation. Note that both IGD and HV can evaluate the convergence and diversity simultaneously. A smaller IGD value or a large HV value indicates a better approximation to the PF. Each algorithm is independently run 51 times. The mean and standard deviation of the IGD and HV values are presented in the corresponding tables, where the ranks of each algorithms on each problems are also given by sorting the mean metric values. The best metric values are highlighted in boldface with a gray background. To have a statistically sound conclusion, we use the Wilcoxon'€s rank sum test at a significant level of 5\% to evaluate whether the proposed MOEA/D-AOOSTM and MOEA/D-AMOSTM are significantly better or worse than the others. In addition, we use the two-sample Kolmogorov-Smirnov test at a significant level of 5\% to summarize the relative performance of all test EMO algorithms.

% !Tex root = main.tex

\section{Empirical Studies}
\label{sec:experiments}

In this section, we first analyze the comparative results for problems with complicated properties. Afterwards, we investigate the effectiveness of the adaptive mechanism. In the end, we summarize the experimental studies in a statistical point of view. Due to the page limits, the empirical studies on problems with more than three objectives are given in the supplementary file of this paper.

\subsection{Performance Comparisons on MOP Instances}

\begin{table*}[!t]
	\scriptsize
	\caption{IGD Results on MOP Test Instances.}
	\label{tab:mop-igd}
	\centering
	% Table generated by Excel2LaTeX from sheet 'v4_final'
	\begin{tabular}{ccccccccccc}
		\toprule
		Problem & IGD   & DRA   & STM   & IR    & AGR   & M2M   & NSGA-III & HypE  & AOOSTM & AMOSTM \\
		\midrule
		& Mean  & 3.380E-1 & 3.509E-1 & 4.726E-2 & 3.189E-2 & \cellcolor[rgb]{0.851, 0.851, 0.851}\textbf{1.614E-2} & 3.652E-1 & 8.013E-1 & 2.407E-2 & 2.390E-2 \\
		MOP1  & Std   & 5.908E-2 & 2.786E-2 & 2.811E-3 & 9.792E-3 & \cellcolor[rgb]{0.851, 0.851, 0.851}\textbf{4.586E-4} & 3.337E-3 & 1.060E-2 & 2.907E-3 & 2.551E-3 \\
		& Rank  & 6 $-$ $\downarrow$ & 7 $-$ $\downarrow$ & 5 $-$ $\downarrow$ & 4 $-$ $\downarrow$ & \cellcolor[rgb]{0.851, 0.851, 0.851}1 $+$ $\uparrow$ & 8 $-$ $\downarrow$ & 9 $-$ $\downarrow$ & 3 $\|$ & 2 \\
		\midrule
		& Mean  & 2.836E-1 & 3.083E-1 & 3.200E-2 & 6.846E-2 & \cellcolor[rgb]{0.851, 0.851, 0.851}\textbf{1.061E-2} & 3.436E-1 & 5.980E-1 & 2.034E-2 & 3.115E-2 \\
		MOP2  & Std   & 7.028E-2 & 6.782E-2 & 2.798E-2 & 7.344E-2 & \cellcolor[rgb]{0.851, 0.851, 0.851}\textbf{1.578E-3} & 1.478E-2 & 2.155E-1 & 4.301E-2 & 6.203E-2 \\
		& Rank  & 6 $-$ $\downarrow$ & 7 $-$ $\downarrow$ & 4 $-$ $\downarrow$ & 5 $-$ $\downarrow$ & \cellcolor[rgb]{0.851, 0.851, 0.851}1 $-$ $\downarrow$ & 8 $-$ $\downarrow$ & 9 $-$ $\downarrow$ & 2 $\|$ & 3 \\
		\midrule
		& Mean  & 4.927E-1 & 4.913E-1 & 4.267E-2 & 6.785E-2 & \cellcolor[rgb]{0.851, 0.851, 0.851}\textbf{1.269E-2} & 3.869E-1 & 6.094E-1 & 4.140E-2 & 3.203E-2 \\
		MOP3  & Std   & 2.885E-2 & 3.391E-2 & 3.691E-2 & 8.518E-2 & \cellcolor[rgb]{0.851, 0.851, 0.851}\textbf{3.924E-3} & 1.337E-16 & 1.742E-1 & 7.378E-2 & 6.527E-2 \\
		& Rank  & 8 $-$ $\downarrow$ & 7 $-$ $\downarrow$ & 4 $-$ $\downarrow$ & 5 $-$ $\downarrow$ & \cellcolor[rgb]{0.851, 0.851, 0.851}1 $-$ $\downarrow$ & 6 $-$ $\downarrow$ & 9 $-$ $\downarrow$ & 3 $\|$ & 2 \\
		\midrule
		& Mean  & 3.068E-1 & 3.136E-1 & 3.843E-2 & 3.934E-2 & \cellcolor[rgb]{0.851, 0.851, 0.851}\textbf{7.774E-3} & 3.147E-1 & 7.107E-1 & 2.025E-2 & 1.414E-2 \\
		MOP4  & Std   & 2.749E-2 & 1.840E-2 & 2.928E-2 & 4.065E-2 & \cellcolor[rgb]{0.851, 0.851, 0.851}\textbf{7.983E-4} & 1.845E-2 & 1.041E-2 & 3.284E-2 & 1.155E-2 \\
		& Rank  & 6 $-$ $\downarrow$ & 7 $-$ $\downarrow$ & 4 $-$ $\downarrow$ & 5 $-$ $\downarrow$ & \cellcolor[rgb]{0.851, 0.851, 0.851}1 $+$ $\uparrow$ & 8 $-$ $\downarrow$ & 9 $-$ $\downarrow$ & 3 $\downarrow$ & 2 \\
		\midrule
		& Mean  & 3.168E-1 & 3.135E-1 & 5.573E-2 & 2.379E-2 & 2.195E-2 & 2.911E-1 & 1.023E+0 & \cellcolor[rgb]{0.851, 0.851, 0.851}\textbf{2.035E-2} & 2.042E-2 \\
		MOP5  & Std   & 7.241E-3 & 1.268E-2 & 2.524E-3 & 3.323E-3 & 2.489E-3 & 2.422E-2 & 2.343E-1 & \cellcolor[rgb]{0.851, 0.851, 0.851}\textbf{1.692E-3} & 1.803E-3 \\
		& Rank  & 8 $-$ $\downarrow$ & 7 $-$ $\downarrow$ & 5 $-$ $\downarrow$ & 4 $-$ $\downarrow$ & 3 $-$ $\downarrow$ & 6 $-$ $\downarrow$ & 9 $-$ $\downarrow$ & \cellcolor[rgb]{0.851, 0.851, 0.851}1 $\|$ & 2 \\
		\midrule
		& Mean  & 3.061E-1 & 3.046E-1 & 1.146E-1 & 8.016E-2 & 8.547E-2 & 3.065E-1 & 5.750E-1 & 5.398E-2 & \cellcolor[rgb]{0.851, 0.851, 0.851}\textbf{5.328E-2} \\
		MOP6  & Std   & 2.161E-8 & 9.552E-3 & 7.590E-3 & 1.015E-2 & 3.941E-3 & 4.459E-4 & 1.620E-2 & 3.094E-3 & \cellcolor[rgb]{0.851, 0.851, 0.851}\textbf{2.917E-3} \\
		& Rank  & 7 $-$ $\downarrow$ & 6 $-$ $\downarrow$ & 5 $-$ $\downarrow$ & 3 $-$ $\downarrow$ & 4 $-$ $\downarrow$ & 8 $-$ $\downarrow$ & 9 $-$ $\downarrow$ & 2 $\|$ & \cellcolor[rgb]{0.851, 0.851, 0.851}1 \\
		\midrule
		& Mean  & 3.501E-1 & 3.512E-1 & 1.778E-1 & 2.458E-1 & 1.171E-1 & 3.514E-1 & 6.377E-1 & 8.186E-2 & \cellcolor[rgb]{0.851, 0.851, 0.851}\textbf{7.912E-2} \\
		MOP7  & Std   & 7.648E-3 & 1.463E-7 & 1.052E-2 & 3.239E-2 & 8.566E-3 & 9.279E-4 & 9.311E-3 & 2.778E-3 & \cellcolor[rgb]{0.851, 0.851, 0.851}\textbf{2.619E-3} \\
		& Rank  & 6 $-$ $\downarrow$ & 7 $-$ $\downarrow$ & 4 $-$ $\downarrow$ & 5 $-$ $\downarrow$ & 3 $-$ $\downarrow$ & 8 $-$ $\downarrow$ & 9 $-$ $\downarrow$ & 2 $\downarrow$ & \cellcolor[rgb]{0.851, 0.851, 0.851}1 \\
		\midrule
		& Total Rank & 47    & 48    & 31    & 31    & 14    & 52    & 63    & 16    & 13 \\
		& Final Rank & 6     & 7     & 4     & 4     & 2     & 8     & 9     & 3     & 1 \\
		\bottomrule
	\end{tabular}%
\begin{tablenotes}
\item[1] According to Wilcoxon's rank sum test, $+$, $-$ and $\approx$ indicate that the corresponding EMO algorithm is significantly better than, worse than or similar to MOEA/D-AOOSTM, while $\uparrow$, $\downarrow$ and $\|$ indicate that the corresponding EMO algorithm is significantly better than, worse than or similar to MOEA/D-AMOSTM.
\end{tablenotes}
\end{table*}

\begin{table*}[!t]
	\scriptsize
	\caption{HV Results on MOP Test Instances.}
	\label{tab:mop-hv}
	\centering
	% Table generated by Excel2LaTeX from sheet 'v4_final'
	\begin{tabular}{ccccccccccc}
		\toprule
		Problem & HV    & DRA   & STM   & IR    & AGR   & M2M   & NSGA-III & HypE  & AOOSTM & AMOSTM \\
		\midrule
		& Mean  & 0.564 & 0.540 & 1.027 & 1.062 & \cellcolor[rgb]{0.851, 0.851, 0.851}\textbf{1.080} & 0.515 & 0.292 & 1.071 & 1.072 \\
		MOP1  & Std   & 1.097E-1 & 5.309E-2 & 4.908E-3 & 1.202E-2 & \cellcolor[rgb]{0.851, 0.851, 0.851}\textbf{9.058E-4} & 8.867E-3 & 1.355E-2 & 3.882E-3 & 3.267E-3 \\
		& Rank  & 6 $-$ $\downarrow$ & 7 $-$ $\downarrow$ & 5 $-$ $\downarrow$ & 4 $-$ $\downarrow$ & \cellcolor[rgb]{0.851, 0.851, 0.851}1 $+$ $\uparrow$ & 8 $-$ $\downarrow$ & 9 $-$ $\downarrow$ & 3 $\|$ & 2 \\
		\midrule
		& Mean  & 0.476 & 0.466 & 0.717 & 0.680 & \cellcolor[rgb]{0.851, 0.851, 0.851}\textbf{0.756} & 0.445 & 0.320 & 0.745 & 0.731 \\
		MOP2  & Std   & 4.459E-2 & 4.257E-2 & 3.322E-2 & 9.570E-2 & \cellcolor[rgb]{0.851, 0.851, 0.851}\textbf{2.340E-3} & 8.938E-3 & 9.798E-2 & 5.037E-2 & 7.825E-2 \\
		& Rank  & 6 $-$ $\downarrow$ & 7 $-$ $\downarrow$ & 4 $-$ $\downarrow$ & 5 $-$ $\downarrow$ & \cellcolor[rgb]{0.851, 0.851, 0.851}1 $-$ $\downarrow$ & 8 $-$ $\downarrow$ & 9 $-$ $\downarrow$ & 2 $\|$ & 3 \\
		\midrule
		& Mean  & 0.240 & 0.240 & 0.595 & 0.560 & \cellcolor[rgb]{0.851, 0.851, 0.851}\textbf{0.637} & 0.440 & 0.316 & 0.606 & 0.617 \\
		MOP3  & Std   & 1.665E-16 & 1.665E-16 & 5.361E-2 & 1.240E-1 & \cellcolor[rgb]{0.851, 0.851, 0.851}\textbf{4.858E-3} & 2.201E-16 & 9.708E-2 & 7.281E-2 & 6.263E-2 \\
		& Rank  & 8 $-$ $\downarrow$ & 9 $-$ $\downarrow$ & 4 $-$ $\downarrow$ & 5 $-$ $\downarrow$ & \cellcolor[rgb]{0.851, 0.851, 0.851}1 $-$ $\downarrow$ & 6 $-$ $\downarrow$ & 7 $-$ $\downarrow$ & 3 $\|$ & 2 \\
		\midrule
		& Mean  & 0.578 & 0.578 & 0.917 & 0.912 & \cellcolor[rgb]{0.851, 0.851, 0.851}\textbf{0.945} & 0.570 & 0.337 & 0.931 & 0.939 \\
		MOP4  & Std   & 2.040E-2 & 1.746E-2 & 4.097E-2 & 5.434E-2 & \cellcolor[rgb]{0.851, 0.851, 0.851}\textbf{2.076E-3} & 9.318E-3 & 1.097E-2 & 4.521E-2 & 1.518E-2 \\
		& Rank  & 6 $-$ $\downarrow$ & 7 $-$ $\downarrow$ & 4 $-$ $\downarrow$ & 5 $-$ $\downarrow$ & \cellcolor[rgb]{0.851, 0.851, 0.851}1 $+$ $\uparrow$ & 8 $-$ $\downarrow$ & 9 $-$ $\downarrow$ & 3 $\downarrow$ & 2 \\
		\midrule
		& Mean  & 0.635 & 0.636 & 1.006 & 1.067 & 1.067 & 0.648 & 0.060 & 1.073 & \cellcolor[rgb]{0.851, 0.851, 0.851}\textbf{1.074} \\
		MOP5  & Std   & 4.447E-9 & 8.201E-3 & 9.637E-3 & 8.135E-3 & 4.295E-3 & 2.991E-2 & 1.806E-1 & 3.038E-3 & \cellcolor[rgb]{0.851, 0.851, 0.851}\textbf{3.196E-3} \\
		& Rank  & 8 $-$ $\downarrow$ & 7 $-$ $\downarrow$ & 5 $-$ $\downarrow$ & 3 $-$ $\downarrow$ & 4 $-$ $\downarrow$ & 6 $-$ $\downarrow$ & 9 $-$ $\downarrow$ & 2 $\|$ & \cellcolor[rgb]{0.851, 0.851, 0.851}1 \\
		\midrule
		& Mean  & 1.221 & 1.224 & 1.418 & 1.463 & 1.439 & 1.216 & 0.682 & 1.494 & \cellcolor[rgb]{0.851, 0.851, 0.851}\textbf{1.495} \\
		MOP6  & Std   & 5.183E-7 & 1.470E-2 & 1.843E-2 & 1.639E-2 & 1.100E-2 & 5.601E-3 & 3.505E-2 & 6.155E-3 & \cellcolor[rgb]{0.851, 0.851, 0.851}\textbf{5.671E-3} \\
		& Rank  & 7 $-$ $\downarrow$ & 6 $-$ $\downarrow$ & 5 $-$ $\downarrow$ & 3 $-$ $\downarrow$ & 4 $-$ $\downarrow$ & 8 $-$ $\downarrow$ & 9 $-$ $\downarrow$ & 2 $\|$ & \cellcolor[rgb]{0.851, 0.851, 0.851}1 \\
		\midrule
		& Mean  & 0.939 & 0.939 & 1.038 & 1.005 & 1.047 & 0.933 & 0.538 & 1.084 & \cellcolor[rgb]{0.851, 0.851, 0.851}\textbf{1.088} \\
		MOP7  & Std   & 2.763E-3 & 1.317E-6 & 2.296E-2 & 4.975E-2 & 2.397E-2 & 5.768E-3 & 6.204E-3 & 5.196E-3 & \cellcolor[rgb]{0.851, 0.851, 0.851}\textbf{4.578E-3} \\
		& Rank  & 6 $-$ $\downarrow$ & 7 $-$ $\downarrow$ & 4 $-$ $\downarrow$ & 5 $-$ $\downarrow$ & 3 $-$ $\downarrow$ & 8 $-$ $\downarrow$ & 9 $-$ $\downarrow$ & 2 $\downarrow$ & \cellcolor[rgb]{0.851, 0.851, 0.851}1 \\
		\midrule
		& Total Rank & 47    & 50    & 31    & 30    & 15    & 52    & 61    & 17    & 12 \\
		& Final Rank & 6     & 7     & 5     & 4     & 2     & 8     & 9     & 3     & 1 \\
		\bottomrule
	\end{tabular}%
\begin{tablenotes}
\item[1] According to Wilcoxon's rank sum test, $+$, $-$ and $\approx$ indicate that the corresponding EMO algorithm is significantly better than, worse than or similar to MOEA/D-AOOSTM, while $\uparrow$, $\downarrow$ and $\|$ indicate that the corresponding EMO algorithm is significantly better than, worse than or similar to MOEA/D-AMOSTM.
\end{tablenotes}
\end{table*}

As discussed in \cite{LiuGZ14}, MOP benchmark suite, in which different parts of the PF have various difficulties, poses significant challenges for maintaining the population diversity. \pref{tab:mop-igd} and \pref{tab:mop-hv} demonstrate the IGD and HV results of the nine EMO algorithms. From the IGD results shown in \pref{tab:mop-igd}, it can be seen that MOEA/D-AMOSTM shows the best overall performance and MOEA/D-AOOSTM, obtaining a slightly lower total rank than MOEA/D-M2M, ranks in the third place. In terms of the mean IGD values, MOEA/D-M2M gives the best results on MOP1 to MOP4, while MOEA/D-AOOSTM ranks the first on MOP5 and MOEA/D-AMOSTM beats all other EMO algorithms on MOP6 and MOP7. When it comes to the Wilcoxon's rank sum test results, both MOEA/D-AMOSTM and MOEA/D-AOOSTM are significantly better the others on MOP2, MOP3 and MOP5 to MOP7. They are only beaten by MOEA/D-M2M on MOP1 and MOP4. This is because MOEA/D-AMOSTM and MOEA/D-AOOSTM achieve better performance on MOP2 and MOP3 than MOEA/D-M2M but the former two have large variances. Comparing MOEA/D-AOOSTM and MOEA/D-AMOSTM, they have no significant differences on five problems but the former is outperformed by the latter on MOP4 and MOP7. Following the best three algorithms, MOEA/D-IR and gMOEA/D-AGR are able to obtain a set of non-dominated solutions moderately covering the entire PF. As for MOEA/D-DRA, MOEA/D-STM, NSGA-III and HypE, they can only obtain some solutions lying on the boundaries. \pref{tab:mop-hv} shows similar results in HV tests, except that MOEA/D-AMOSTM obtains better performance than MOEA/D-AOOSTM on MOP5. 

We plot the final solution sets with the best IGD values in 51 runs on all test instances in the supplementary file of this paper. From Fig. 1 to Fig. 4 of the supplementary file, we can see that although MOEA/D-M2M obtains slightly better mean IGD and HV metric values than MOEA/D-AMOSTM and MOEA/D-AOOSTM, the solutions obtained by MOEA/D-AMOSTM and MOEA/D-AOOSTM have a more uniform distribution along the PF. This can be explained by the density estimation method, i.e., the crowding distance of NSGA-II, used in MOEA/D-M2M, which is too coarse to guarantee the population diversity. Nevertheless, the convergence ability of MOEA/D-M2M is satisfied, thus contributing to promising IGD values on MOP1 to MOP4. According to \cite{WangZZGJ16}, gMOEA/D-AGR uses a sigmoid function to assign a same replacement neighborhood size to all subproblems. However, since different parts of the PF require various efforts, this same setting might not be appropriate for all subproblems. From Fig.~3 and Fig.~7 of the supplementary file, we can obverse that the solutions obtained by gMOEA/D-AGR may miss some segments of the PF. This can be explained by the replacement neighborhood that grows too fast for the corresponding subproblems. In order to emphasize the population diversity, for each subproblem, MOEA/D-IR selects the appropriate solution from a couple of related ones. However, its preference setting, which encourages the selection in a less crowded area, tends to result in an unstable selection result. In this case, some solutions far away from the PF can be selected occasionally. The reason behind the poor performance of NSGA-III, HypE and MOEA/D-DRA is that their convergence first and diversity second selection strategies may easily trap the population in some narrow areas. As discussed in \pref{sec:drawbacks}, the stable matching model used in MOEA/D-STM can easily match a solution with an unfavorable subproblem, thus resulting in an unbalanced selection.

\subsection{Performance Comparisons on UF Instances}

\begin{table*}[!t]
	\scriptsize
	\caption{IGD Results on UF Test Instances.}
	\label{tab:uf-igd}
	\centering
	% Table generated by Excel2LaTeX from sheet 'v4_final'
	\begin{tabular}{ccccccccccc}
		\toprule
		Problem & IGD   & DRA   & STM   & IR    & AGR   & M2M   & NSGA-III & HypE  & AOOSTM & AMOSTM \\
		\midrule
		& Mean  & 1.071E-3 & 1.043E-3 & 2.471E-3 & 1.813E-3 & 7.076E-3 & 9.457E-2 & 9.902E-2 & \cellcolor[rgb]{0.851, 0.851, 0.851}\textbf{9.631E-4} & 9.696E-4 \\
		UF1   & Std   & 2.583E-4 & 7.870E-5 & 1.180E-4 & 8.699E-5 & 2.785E-3 & 1.200E-2 & 1.089E-2 & \cellcolor[rgb]{0.851, 0.851, 0.851}\textbf{4.650E-5} & 5.158E-5 \\
		& Rank  & 4 $-$ $\downarrow$ & 3 $-$ $\downarrow$ & 6 $-$ $\downarrow$ & 5 $-$ $\downarrow$ & 7 $-$ $\downarrow$ & 8 $-$ $\downarrow$ & 9 $-$ $\downarrow$ & \cellcolor[rgb]{0.851, 0.851, 0.851}1 $\|$ & 2 \\
		\midrule
		& Mean  & 4.601E-3 & 3.024E-3 & 5.475E-3 & 5.256E-3 & 3.957E-3 & 2.993E-2 & 2.119E-1 & \cellcolor[rgb]{0.851, 0.851, 0.851}\textbf{2.270E-3} & 2.577E-3 \\
		UF2   & Std   & 9.338E-3 & 9.309E-4 & 1.172E-3 & 7.183E-4 & 5.099E-4 & 2.629E-3 & 6.301E-2 & \cellcolor[rgb]{0.851, 0.851, 0.851}\textbf{5.587E-4} & 5.649E-4 \\
		& Rank  & 5 $\approx$ $\|$ & 3 $-$ $\downarrow$ & 7 $-$ $\downarrow$ & 6 $-$ $\downarrow$ & 4 $-$ $\downarrow$ & 8 $-$ $\downarrow$ & 9 $-$ $\downarrow$ & \cellcolor[rgb]{0.851, 0.851, 0.851}1 $\uparrow$ & 2 \\
		\midrule
		& Mean  & 1.772E-2 & 7.757E-3 & 1.642E-2 & 8.141E-3 & 1.549E-2 & 2.078E-1 & 1.805E-1 & 7.296E-3 & \cellcolor[rgb]{0.851, 0.851, 0.851}\textbf{4.110E-3} \\
		UF3   & Std   & 1.500E-2 & 6.213E-3 & 1.289E-2 & 8.673E-3 & 5.495E-3 & 4.775E-2 & 5.100E-2 & 8.380E-3 & \cellcolor[rgb]{0.851, 0.851, 0.851}\textbf{3.128E-3} \\
		& Rank  & 7 $-$ $\downarrow$ & 3 $\approx$ $\downarrow$ & 6 $-$ $\downarrow$ & 4 $\approx$ $\downarrow$ & 5 $-$ $\downarrow$ & 9 $-$ $\downarrow$ & 8 $-$ $\downarrow$ & 2 $\|$ & \cellcolor[rgb]{0.851, 0.851, 0.851}1 \\
		\midrule
		& Mean  & 5.320E-2 & 5.076E-2 & 5.623E-2 & 5.025E-2 & \cellcolor[rgb]{0.851, 0.851, 0.851}\textbf{3.994E-2} & 4.297E-2 & 4.899E-2 & 5.269E-2 & 5.043E-2 \\
		UF4   & Std   & 3.115E-3 & 2.857E-3 & 2.818E-3 & 2.874E-3 & \cellcolor[rgb]{0.851, 0.851, 0.851}\textbf{3.705E-4} & 8.311E-4 & 7.077E-3 & 3.523E-3 & 2.803E-3 \\
		& Rank  & 8 $\approx$ $\downarrow$ & 6 $+$ $\|$ & 9 $-$ $\downarrow$ & 4 $+$ $\|$ & \cellcolor[rgb]{0.851, 0.851, 0.851}1 $+$ $\uparrow$ & 2 $+$ $\uparrow$ & 3 $+$ $\uparrow$ & 7 $\downarrow$ & 5 \\
		\midrule
		& Mean  & 3.033E-1 & 2.397E-1 & 2.574E-1 & 2.625E-1 & \cellcolor[rgb]{0.851, 0.851, 0.851}\textbf{1.795E-1} & 2.107E-1 & 2.289E-1 & 2.514E-1 & 2.392E-1 \\
		UF5   & Std   & 7.779E-2 & 3.369E-2 & 4.334E-2 & 1.102E-1 & \cellcolor[rgb]{0.851, 0.851, 0.851}\textbf{3.013E-2} & 2.131E-2 & 4.852E-2 & 1.766E-2 & 2.220E-2 \\
		& Rank  & 9 $-$ $\downarrow$ & 5 $+$ $\|$ & 7 $\approx$ $\downarrow$ & 8 $\approx$ $\|$ & \cellcolor[rgb]{0.851, 0.851, 0.851}1 $+$ $\uparrow$ & 2 $+$ $\uparrow$ & 3 $+$ $\uparrow$ & 6 $\downarrow$ & 4 \\
		\midrule
		& Mean  & 1.504E-1 & 7.805E-2 & 1.073E-1 & 1.126E-1 & 8.990E-2 & 2.134E-1 & 2.312E-1 & 8.146E-2 & \cellcolor[rgb]{0.851, 0.851, 0.851}\textbf{6.876E-2} \\
		UF6   & Std   & 1.224E-1 & 4.305E-2 & 4.600E-2 & 7.840E-2 & 5.355E-2 & 6.523E-2 & 6.828E-2 & 4.048E-2 & \cellcolor[rgb]{0.851, 0.851, 0.851}\textbf{3.300E-2} \\
		& Rank  & 7 $-$ $\downarrow$ & 2 $+$ $\|$ & 5 $-$ $\downarrow$ & 6 $-$ $\downarrow$ & 4 $\approx$ $\downarrow$ & 8 $-$ $\downarrow$ & 9 $-$ $\downarrow$ & 3 $\downarrow$ & \cellcolor[rgb]{0.851, 0.851, 0.851}1 \\
		\midrule
		& Mean  & 1.245E-3 & \cellcolor[rgb]{0.851, 0.851, 0.851}\textbf{1.123E-3} & 3.707E-3 & 2.145E-3 & 6.234E-3 & 6.856E-2 & 2.622E-1 & 1.150E-3 & 1.148E-3 \\
		UF7   & Std   & 2.371E-4 & \cellcolor[rgb]{0.851, 0.851, 0.851}\textbf{7.371E-5} & 5.295E-4 & 3.221E-4 & 1.867E-3 & 8.357E-2 & 4.540E-2 & 1.095E-4 & 1.481E-4 \\
		& Rank  & 4 $\approx$ $\downarrow$ & \cellcolor[rgb]{0.851, 0.851, 0.851}1 $\approx$ $\|$ & 6 $-$ $\downarrow$ & 5 $-$ $\downarrow$ & 7 $-$ $\downarrow$ & 8 $-$ $\downarrow$ & 9 $-$ $\downarrow$ & 3 $\|$ & 2 \\
		\midrule
		& Mean  & 3.104E-2 & 3.019E-2 & 6.467E-2 & 4.715E-2 & 9.655E-2 & 1.674E-1 & 3.116E-1 & \cellcolor[rgb]{0.851, 0.851, 0.851}\textbf{2.921E-2} & 5.393E-2 \\
		UF8   & Std   & 4.020E-3 & 8.706E-3 & 1.070E-2 & 9.477E-3 & 8.181E-3 & 2.670E-3 & 3.417E-2 & \cellcolor[rgb]{0.851, 0.851, 0.851}\textbf{5.154E-3} & 9.528E-3 \\
		& Rank  & 3 $-$ $\uparrow$ & 2 $\approx$ $\uparrow$ & 6 $-$ $\downarrow$ & 4 $-$ $\uparrow$ & 7 $-$ $\downarrow$ & 8 $-$ $\downarrow$ & 9 $-$ $\downarrow$ & \cellcolor[rgb]{0.851, 0.851, 0.851}1 $\uparrow$ & 5 \\
		\midrule
		& Mean  & 4.779E-2 & \cellcolor[rgb]{0.851, 0.851, 0.851}\textbf{2.373E-2} & 5.794E-2 & 5.861E-2 & 1.148E-1 & 1.767E-1 & 2.353E-1 & 3.704E-2 & 3.769E-2 \\
		UF9   & Std   & 3.446E-2 & \cellcolor[rgb]{0.851, 0.851, 0.851}\textbf{1.112E-3} & 3.960E-2 & 4.572E-2 & 3.045E-2 & 3.924E-2 & 3.018E-2 & 3.125E-2 & 4.306E-2 \\
		& Rank  & 4 $-$ $\downarrow$ & \cellcolor[rgb]{0.851, 0.851, 0.851}1 $+$ $\|$ & 5 $-$ $\downarrow$ & 6 $-$ $\downarrow$ & 7 $-$ $\downarrow$ & 8 $-$ $\downarrow$ & 9 $-$ $\downarrow$ & 2 $\downarrow$ & 3 \\
		\midrule
		& Mean  & 5.184E-1 & 1.701E+0 & 7.216E-1 & 4.168E-1 & 5.572E-1 & \cellcolor[rgb]{0.851, 0.851, 0.851}\textbf{2.257E-1} & 2.568E-1 & 1.028E+0 & 2.426E+0 \\
		UF10  & Std   & 6.698E-2 & 2.849E-1 & 1.202E-1 & 7.165E-2 & 5.950E-2 & \cellcolor[rgb]{0.851, 0.851, 0.851}\textbf{5.700E-2} & 6.938E-2 & 2.943E-1 & 1.868E-1 \\
		& Rank  & 4 $+$ $\uparrow$ & 8 $-$ $\uparrow$ & 6 $+$ $\uparrow$ & 3 $+$ $\uparrow$ & 5 $+$ $\uparrow$ & \cellcolor[rgb]{0.851, 0.851, 0.851}1 $+$ $\uparrow$ & 2 $+$ $\uparrow$ & 7 $\uparrow$ & 9 \\
		\midrule
		& Total Rank & 55    & 34    & 63    & 51    & 48    & 62    & 70    & 33    & 34 \\
		& Final Rank & 6     & 2     & 8     & 5     & 4     & 7     & 9     & 1     & 2 \\
		\bottomrule
	\end{tabular}%
	\begin{tablenotes}
		\item[1] According to Wilcoxon's rank sum test, $+$, $-$ and $\approx$ indicate that the corresponding EMO algorithm is significantly better than, worse than or similar to MOEA/D-AOOSTM, while $\uparrow$, $\downarrow$ and $\|$ indicate that the corresponding EMO algorithm is significantly better than, worse than or similar to MOEA/D-AMOSTM.
	\end{tablenotes}
\end{table*}

\begin{table*}[!t]
	\scriptsize
	\caption{HV Results on UF Test Instances.}
	\label{tab:uf-hv}
	\centering
	% Table generated by Excel2LaTeX from sheet 'v4_final'
	\begin{tabular}{ccccccccccc}
		\toprule
		Problem & HV    & DRA   & STM   & IR    & AGR   & M2M   & NSGA-III & HypE  & AOOSTM & AMOSTM \\
		\midrule
		& Mean  & 1.104 & 1.104 & 1.101 & 1.102 & 1.092 & 0.945 & 0.941 & \cellcolor[rgb]{0.851, 0.851, 0.851}\textbf{1.104} & 1.104 \\
		UF1   & Std   & 6.732E-4 & 4.643E-4 & 6.668E-4 & 4.079E-4 & 5.167E-3 & 2.713E-2 & 2.800E-2 & \cellcolor[rgb]{0.851, 0.851, 0.851}\textbf{3.180E-4} & 4.627E-4 \\
		& Rank  & 4 $-$ $\|$ & 3 $-$ $\|$ & 6 $-$ $\downarrow$ & 5 $-$ $\downarrow$ & 7 $-$ $\downarrow$ & 8 $-$ $\downarrow$ & 9 $-$ $\downarrow$ & \cellcolor[rgb]{0.851, 0.851, 0.851}1 $\uparrow$ & 2 \\
		\midrule
		& Mean  & 1.097 & 1.100 & 1.093 & 1.096 & 1.099 & 1.054 & 0.889 & \cellcolor[rgb]{0.851, 0.851, 0.851}\textbf{1.101} & 1.101 \\
		UF2   & Std   & 1.238E-2 & 1.889E-3 & 3.835E-3 & 1.866E-3 & 1.903E-3 & 6.402E-3 & 4.580E-2 & \cellcolor[rgb]{0.851, 0.851, 0.851}\textbf{1.730E-3} & 1.395E-3 \\
		& Rank  & 5 $-$ $\|$ & 3 $-$ $\|$ & 7 $-$ $\downarrow$ & 6 $-$ $\downarrow$ & 4 $-$ $\downarrow$ & 8 $-$ $\downarrow$ & 9 $-$ $\downarrow$ & \cellcolor[rgb]{0.851, 0.851, 0.851}1 $\uparrow$ & 2 \\
		\midrule
		& Mean  & 1.073 & 1.093 & 1.075 & 1.090 & 1.079 & 0.732 & 0.793 & 1.094 & \cellcolor[rgb]{0.851, 0.851, 0.851}\textbf{1.099} \\
		UF3   & Std   & 2.867E-2 & 1.066E-2 & 2.590E-2 & 1.848E-2 & 8.270E-3 & 5.290E-2 & 6.111E-2 & 1.513E-2 & \cellcolor[rgb]{0.851, 0.851, 0.851}\textbf{5.436E-3} \\
		& Rank  & 7 $-$ $\downarrow$ & 3 $-$ $\downarrow$ & 6 $-$ $\downarrow$ & 4 $\approx$ $\downarrow$ & 5 $-$ $\downarrow$ & 9 $-$ $\downarrow$ & 8 $-$ $\downarrow$ & 2 $\|$ & \cellcolor[rgb]{0.851, 0.851, 0.851}1 \\
		\midrule
		& Mean  & 0.672 & 0.679 & 0.667 & 0.680 & \cellcolor[rgb]{0.851, 0.851, 0.851}\textbf{0.701} & 0.698 & 0.685 & 0.676 & 0.679 \\
		UF4   & Std   & 5.996E-3 & 5.759E-3 & 5.444E-3 & 5.213E-3 & \cellcolor[rgb]{0.851, 0.851, 0.851}\textbf{7.356E-4} & 1.257E-3 & 1.459E-2 & 6.500E-3 & 5.755E-3 \\
		& Rank  & 8 $-$ $\downarrow$ & 6 $\approx$ $\|$ & 9 $-$ $\downarrow$ & 4 $+$ $\|$ & \cellcolor[rgb]{0.851, 0.851, 0.851}1 $+$ $\uparrow$ & 2 $+$ $\uparrow$ & 3 $+$ $\uparrow$ & 7 $\downarrow$ & 5 \\
		\midrule
		& Mean  & 0.353 & 0.437 & 0.414 & 0.455 & \cellcolor[rgb]{0.851, 0.851, 0.851}\textbf{0.574} & 0.536 & 0.519 & 0.411 & 0.434 \\
		UF5   & Std   & 8.320E-2 & 7.346E-2 & 8.085E-2 & 1.201E-1 & \cellcolor[rgb]{0.851, 0.851, 0.851}\textbf{6.204E-2} & 3.906E-2 & 7.893E-2 & 3.560E-2 & 4.814E-2 \\
		& Rank  & 9 $-$ $\downarrow$ & 5 $+$ $\|$ & 7 $\approx$ $\|$ & 4 $+$ $\|$ & \cellcolor[rgb]{0.851, 0.851, 0.851}1 $+$ $\uparrow$ & 2 $+$ $\uparrow$ & 3 $+$ $\uparrow$ & 8 $\downarrow$ & 6 \\
		\midrule
		& Mean  & 0.591 & 0.645 & 0.610 & 0.647 & \cellcolor[rgb]{0.851, 0.851, 0.851}\textbf{0.685} & 0.621 & 0.592 & 0.646 & 0.666 \\
		UF6   & Std   & 1.152E-1 & 8.742E-2 & 8.839E-2 & 6.045E-2 & \cellcolor[rgb]{0.851, 0.851, 0.851}\textbf{4.863E-2} & 2.851E-2 & 6.576E-2 & 8.337E-2 & 6.562E-2 \\
		& Rank  & 9 $-$ $\downarrow$ & 5 $\approx$ $\|$ & 7 $-$ $\downarrow$ & 3 $\approx$ $\downarrow$ & \cellcolor[rgb]{0.851, 0.851, 0.851}1 $\approx$ $\|$ & 6 $-$ $\downarrow$ & 8 $-$ $\downarrow$ & 4 $\downarrow$ & 2 \\
		\midrule
		& Mean  & 0.937 & \cellcolor[rgb]{0.851, 0.851, 0.851}\textbf{0.937} & 0.931 & 0.935 & 0.928 & 0.831 & 0.610 & 0.937 & 0.937 \\
		UF7   & Std   & 9.245E-4 & \cellcolor[rgb]{0.851, 0.851, 0.851}\textbf{4.244E-4} & 1.589E-3 & 1.674E-3 & 3.597E-3 & 1.055E-1 & 3.176E-2 & 6.227E-4 & 6.386E-4 \\
		& Rank  & 4 $-$ $\|$ & \cellcolor[rgb]{0.851, 0.851, 0.851}1 $\approx$ $\uparrow$ & 6 $-$ $\downarrow$ & 5 $-$ $\downarrow$ & 7 $-$ $\downarrow$ & 8 $-$ $\downarrow$ & 9 $-$ $\downarrow$ & 2 $\uparrow$ & 3 \\
		\midrule
		& Mean  & 1.127 & 1.125 & 1.050 & 1.088 & 0.938 & 0.777 & 0.783 & \cellcolor[rgb]{0.851, 0.851, 0.851}\textbf{1.143} & 1.073 \\
		UF8   & Std   & 9.272E-3 & 1.428E-2 & 2.737E-2 & 2.234E-2 & 2.383E-2 & 4.349E-3 & 3.482E-3 & \cellcolor[rgb]{0.851, 0.851, 0.851}\textbf{1.291E-2} & 2.411E-2 \\
		& Rank  & 2 $-$ $\uparrow$ & 3 $-$ $\uparrow$ & 6 $-$ $\downarrow$ & 4 $-$ $\uparrow$ & 7 $-$ $\downarrow$ & 9 $-$ $\downarrow$ & 8 $-$ $\downarrow$ & \cellcolor[rgb]{0.851, 0.851, 0.851}1 $\uparrow$ & 5 \\
		\midrule
		& Mean  & 1.402 & \cellcolor[rgb]{0.851, 0.851, 0.851}\textbf{1.462} & 1.400 & 1.395 & 1.243 & 0.971 & 0.855 & 1.455 & 1.453 \\
		UF9   & Std   & 6.489E-2 & \cellcolor[rgb]{0.851, 0.851, 0.851}\textbf{3.831E-3} & 7.704E-2 & 9.007E-2 & 5.711E-2 & 7.193E-2 & 8.419E-2 & 6.385E-2 & 8.895E-2 \\
		& Rank  & 4 $-$ $\downarrow$ & \cellcolor[rgb]{0.851, 0.851, 0.851}1 $-$ $\downarrow$ & 5 $-$ $\downarrow$ & 6 $-$ $\downarrow$ & 7 $-$ $\downarrow$ & 8 $-$ $\downarrow$ & 9 $-$ $\downarrow$ & 2 $\downarrow$ & 3 \\
		\midrule
		& Mean  & 0.188 & 0.000 & 0.063 & 0.311 & 0.172 & \cellcolor[rgb]{0.851, 0.851, 0.851}\textbf{0.653} & 0.612 & 0.015 & 0.000 \\
		UF10  & Std   & 4.694E-2 & 3.781E-4 & 4.914E-2 & 6.402E-2 & 3.272E-2 & \cellcolor[rgb]{0.851, 0.851, 0.851}\textbf{1.121E-1} & 1.286E-1 & 3.031E-2 & 0.000E+0 \\
		& Rank  & 4 $+$ $\uparrow$ & 8 $-$ $\|$ & 6 $+$ $\uparrow$ & 3 $+$ $\uparrow$ & 5 $+$ $\uparrow$ & \cellcolor[rgb]{0.851, 0.851, 0.851}1 $+$ $\uparrow$ & 2 $+$ $\uparrow$ & 7 $\uparrow$ & 9 \\
		\midrule
		& Total Rank & 56    & 38    & 65    & 44    & 45    & 61    & 68    & 35    & 38 \\
		& Final Rank & 6     & 2     & 8     & 4     & 5     & 7     & 9     & 1     & 2 \\
		\bottomrule
	\end{tabular}%
\begin{tablenotes}
\item[1] According to Wilcoxon's rank sum test, $+$, $-$ and $\approx$ indicate that the corresponding EMO algorithm is significantly better than, worse than or similar to MOEA/D-AOOSTM, while $\uparrow$, $\downarrow$ and $\|$ indicate that the corresponding EMO algorithm is significantly better than, worse than or similar to MOEA/D-AMOSTM.
\end{tablenotes}
\end{table*}

The comparison results on the IGD and HV metrics between MOEA/D-AOOSTM, MOEA/D-AMOSTM and the other EMO algorithms on UF benchmark suite are presented in \pref{tab:uf-igd} and \pref{tab:uf-hv}. Different from the MOP benchmark suite, the major source of difficulty for the UF benchmark suite is not the diversity preservation but the complicated PS. Generally speaking, the overall performance of MOEA/D-AOOSTM ranks the first on the UF benchmark suite, followed by MOEA/D-AMOSTM and their predecessor MOEA/D-STM. More specifically, for both the IGD and HV metrics, MOEA/D-AOOSTM performs the best on UF1, UF2 and UF8 and acts as the top three algorithm on all instances except for UF4, UF5 and UF10. For UF3 and UF7, the performance of MOEA/D-AOOSTM does not show significant difference with the best performing algorithms. MOEA/D-AMOSTM shows similar rankings to MOEA/D-AOOSTM. It is significantly better than MOEA/D-AOOSTM on UF4-UF6 and UF9 in terms of both the IGD and HV metrics. In contrast, MOEA/D-AOOSTM wins on UF2, UF8 and UF10 according to Wilcoxon's rank sum test of the IGD results and wins on UF1, UF2, UF7, UF8 and UF10 in the HV tests. 

According to the performance of different algorithms on the UF test instances, the analysis can be divided into three groups. For UF4 and UF5, MOEA/D-M2M, NSGA-III and HypE are able to provide better performance than all other MOEA/D variants. All these three algorithms use the Pareto dominance as the major driving force in the environmental selection, which can improve the convergence to a great extent. For UF1 to UF3 and UF6 to UF9, all the MOEA/D variants outperform NSGA-III and HypE. In particular, The three variants with stable matching-based selection, i.e., MOEA/D-AOOSTM, MOEA/D-AMOSTM and MOEA/D-STM, have shown very promising results on these six test instances. The superior performance can be attributed to the well balance between convergence and diversity achieved by the stable matching relationship between subproblems and solutions. gMOEA/D-AGR has shown a medium performance for the former two groups of problem instances. This might be due to its adaptive mechanism that can hardly make a satisfied prediction of the replacement neighborhood size. UF10 is a difficult tri-objective problem, where none of these eight EMO algorithms are able to obtain a well approximation to the PF within the given number of function evaluations. Nevertheless, it is worth noting that the empirical studies in~\cite{LiZKLW14} demonstrate that the stable matching-based selection mechanism can offer a competitive result in case the maximum number of function evaluations is doubled.

\subsection{Performance Comparisons on Bi-Objective WFG Instances}

From the comparison results shown in Table I and Table II of the supplementary file, it can be seen that MOEA/D-AOOSTM and MOEA/D-AMOSTM are the best two algorithms in overall performance on the bi-objective WFG instances. Comparing with the seven existing algorithms, MOEA/D-AOOSTM and MOEA/D-AMOSTM achieves significant better performance in 56 and 57 out of 63 IGD comparisons respectively. As for HV results, they both wins in 57 comparisons. In particular, MOEA/D-AOOSTM and MOEA/D-AMOSTM obtain the best mean metric values on WFG1, WFG3, WFG6, WFG7 and WFG9 and obtain very promising results on WFG3 and WFG5. Even though the mean HV metric values of MOEA/D-AOOSTM and MOEA/D-AMOSTM on WFG2 rank the fifth and sixth, all the other algorithms are significantly worse than them. It is interesting to note that MOEA/D-AOOSTM and MOEA/D-AMOSTM are significantly better than MOEA/D-STM on all WFG instances except for WFG8. One possible reason is the proper normalization method used in MOEA/D-AOOSTM/AMOSTM. However, we also notice that the performance of NSGA-III fluctuates significantly on difficult problem instances, though it uses the same normalization method. The other variants of MOEA/D perform more or less the same on all test instances except MOEA/D-M2M significantly outperforms all other EMO algorithms on WFG8. The indicator-based algorithm HypE performs the worst on all nine problems. Comparing the IGD and HV results, the algorithm comparisons are consistent on most test instances, except when the performance of two algorithms are very close, the ranking of IGD and HV metric values may change slightly. However, it is worth noting that the algorithms perform quite differently on WFG2 under the IGD and HV assessments. NSGA-III shows the best mean IGD value but gives the second worst mean HV value. In contrast, the mean HV value of gMOEA/D-AGR ranks the first among all algorithms but its mean IGD value only obtains a rank of $7$. This is probably because WFG2 has a discontinuous PF, which makes the distinction between IGD and HV more obvious.

%\vspace{-1.0em}

\subsection{Effectiveness of the Adaptive Mechanism}
\label{sec:adaptive-validate}

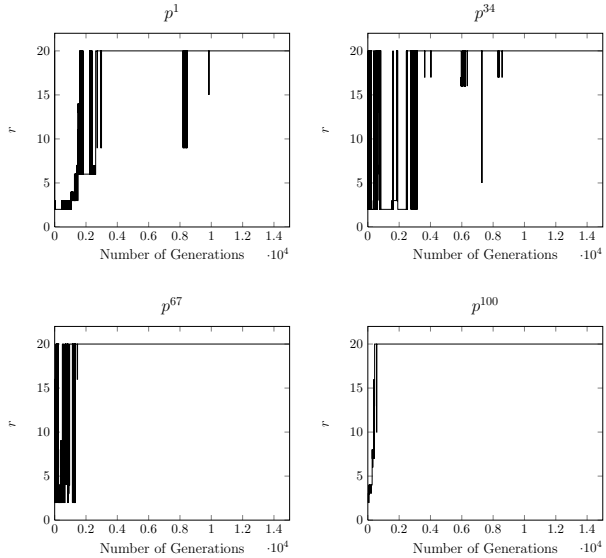
\begin{figure}[!t]
	\centering
	\subfloat{
		\resizebox{0.23\textwidth}{!}{
			\begin{tikzpicture}
			\begin{axis}[
			title  = $p^1$,
			xmin   = 0, xmax = 15000,
			ymin   = 0, ymax = 22,
			label style = {font=\large},
			title style={font=\Large},
			xlabel = {Number of Generations},
			ylabel = {$r$}
			]
			\addplot[black, samples y = 0] table[x index = 0, y index = 1] {Data/rplot_p1.dat};
			\end{axis}
			\end{tikzpicture}
		}
	}
	\subfloat{
		\resizebox{0.23\textwidth}{!}{
			\begin{tikzpicture}
			\begin{axis}[
			title  = $p^{34}$,
			xmin   = 0, xmax = 15000,
			ymin   = 0, ymax = 22,
			label style = {font=\large},
			title style={font=\Large},
			xlabel = {Number of Generations},
			ylabel = {$r$}
			]
			\addplot[black, samples y = 0] table[x index = 0, y index = 1] {Data/rplot_p34.dat};
			\end{axis}
			\end{tikzpicture}
		}
	}\\
	\subfloat{
		\resizebox{0.23\textwidth}{!}{
			\begin{tikzpicture}
			\begin{axis}[
			title  = $p^{67}$,
			xmin   = 0, xmax = 15000,
			ymin   = 0, ymax = 22,
			label style = {font=\large},
			title style={font=\Large},
			xlabel = {Number of Generations},
			ylabel = {$r$}
			]
			\addplot[black, samples y = 0] table[x index = 0, y index = 1] {Data/rplot_p67.dat};
			\end{axis}
			\end{tikzpicture}
		}
	}
	\subfloat{
		\resizebox{0.23\textwidth}{!}{
			\begin{tikzpicture}
			\begin{axis}[
			title  = $p^{100}$,
			xmin   = 0, xmax = 15000,
			ymin   = 0, ymax = 22,
			label style = {font=\large},
			title style={font=\Large},
			xlabel = {Number of Generations},
			ylabel = {$r$}
			]
			\addplot[black, samples y = 0] table[x index = 0, y index = 1] {Data/rplot_p100.dat};
			\end{axis}
			\end{tikzpicture}
		}
	}
	\caption{The illustration of the $r$ values of the solutions selected by subproblems $p_1$, $p_{34}$, $p_{67}$ and $p_{100}$.}
	\label{fig:subp-r}
\end{figure}

To show the functionality of the adaptive mechanism proposed in \pref{sec:adaptive}, we choose MOP1 as an example and plot the trajectories of $r$ of the selected solutions of four different subproblems, i.e., $p^1$, $p^{34}$, $p^{67}$ and $p^{100}$, controlled by the local competitiveness-based adaptive mechanism. From trajectories shown in \pref{fig:subp-r}, we notice that the $r$ value fluctuates significantly at the early stages of the evolution. Afterwards, it almost convergences to the threshold $\ell_{max}$. This is because the local competitiveness varies dramatically when the population is far away from but heading to the PF. With the progress of evolution, the selected solutions gradually become non-dominated from each other. As a consequence, the value of $\ell$, which keeps the solution locally noncompetitive, grows and finally settles at $\ell_{max}$. All in all, we can see that different solutions have different local competitiveness, thus it is meaningful to have a different length of the preference list.

In order to further investigate the effectiveness brought by our proposed adaptive mechanism, we develop two variants, denoted by MOEA/D-AOOSTM-$v$ and MOEA/D-AMOSTM-$v$, in which all solutions share the same static $r$ setting. From the our offline parameter studies, we finally find that $r=4$ and $r=8$ are the best settings for MOP and UF benchmark suites respectively. Thereafter, in Table VII of the supplementary file, we show the IGD results of MOEA/D-AOOSTM-$v$ and MOEA/D-AMOSTM-$v$ with the best $r$ settings. Comparing with the best static settings of $r$, we can see that both MOEA/D-AOOSTM and MOEA/D-AMOSTM obtain better mean IGD values than their variant on 11 out of 17 problems. Even though MOEA/D-AOOSTM-$v$ and MOEA/D-AMOSTM-$v$ perform better on some test instances, in most of the cases, our proposed adaptive mechanism achieve comparable results to the best settings of $r$. Note that the optimal $r$ settings are obtained from a series of comprehensive try-and-error experiments, which are not as intelligent and flexible as our proposed adaptive mechanism. Therefore, we conclude that our proposed adaptive mechanism based on the local competitiveness is generally effective for dynamically setting the length of the solution's preference list.

\subsection{Summaries}

\begin{table*}[!t]
	\scriptsize
	\caption{Final Ranks of Mean Metric Values on MOP, UF and Bi-Objective WFG Test Instances.}
	\label{tab:rank-multi}
	\centering
	% Table generated by Excel2LaTeX from sheet 'v4_final'
	\begin{tabular}{ccccccccccc}
		\toprule
		Metric & Rank & ~~DRA~~   & ~~STM~~   & ~~~IR~~~    & ~~AGR~~   & ~~M2M~~   & NSGA-III & ~~HypE~~  & AOOSTM & AMOSTM \\
		\midrule
		\multirow{2}[2]{*}{IGD} & Total Rank & 146   & 124   & 160   & 123   & 115   & 154   & 214   & 68    & 66 \\
		& Final Rank & 6     & 5     & 8     & 4     & 3     & 7     & 9     & 2     & 1 \\
		\midrule
		\multirow{2}[2]{*}{HV} & Total Rank & 147   & 132   & 165   & 104   & 109   & 157   & 209   & 72    & 75 \\
		& Final Rank & 6     & 5     & 8     & 3     & 4     & 7     & 9     & 1     & 2 \\
		\bottomrule
	\end{tabular}%
\end{table*}

\begin{table*}[!t]
	\scriptsize
	\caption{Final Ranks of Mean Metric Values on 3-, 5-, 8- and 10-Objective WFG Test Instances.}
	\label{tab:rank-many}
	\centering
	% Table generated by Excel2LaTeX from sheet 'v4_final'
	\begin{tabular}{ccccccccc}
		\toprule
		Metric & Rank & ~~DRA~~   & MOEA/DD & PICEA-g & NSGA-III & ~~HypE~~  & AOOSTM & AMOSTM \\
		\midrule
		\multirow{2}[2]{*}{HV} & Total Rank & 223   & 153   & 133   & 120   & 226   & 70    & 83 \\
		& Final Rank & 6     & 5     & 4     & 3     & 7     & 1     & 2 \\
		\bottomrule
	\end{tabular}%
\end{table*}

\begin{table*}[!t]
	\scriptsize
	\caption{Average Performance Scores on MOP, UF and Bi-Objective WFG Test Instances.}
	\label{tab:score-multi}
	\centering
	% Table generated by Excel2LaTeX from sheet 'v4_final'
	\begin{tabular}{ccccccccccc}
		\toprule
		Metric & Problem & ~~DRA~~   & ~~STM~~   & ~~~IR~~~    & ~~AGR~~   & ~~M2M~~   & NSGA-III & ~~HypE~~  & AOOSTM & AMOSTM \\
		\midrule
		\multirow{4}[2]{*}{IGD} & MOP   & 6.14  & 5.86  & 3.86  & 3.43  & 1.57  & 6.43  & 8.00  & \cellcolor[rgb]{0.851, 0.851, 0.851} 0.57 & \cellcolor[rgb]{0.851, 0.851, 0.851} \textbf{0.29} \\
		& UF    & 3.90  & 2.30  & 5.40  & 3.30  & 4.00  & 5.20  & 5.90  & \cellcolor[rgb]{0.851, 0.851, 0.851} 2.30 & \cellcolor[rgb]{0.851, 0.851, 0.851} \textbf{2.10} \\
		& WFG   & 3.44  & 3.78  & 6.78  & 3.56  & 5.44  & 3.33  & 7.89  & \cellcolor[rgb]{0.851, 0.851, 0.851} 1.22 & \cellcolor[rgb]{0.851, 0.851, 0.851} \textbf{0.67} \\
		& Overall & 4.35  & 3.77  & 5.46  & 3.42  & 3.85  & 4.88  & 7.15  & \cellcolor[rgb]{0.851, 0.851, 0.851} 1.46 & \cellcolor[rgb]{0.851, 0.851, 0.851} \textbf{1.12} \\
		\midrule
		\multirow{4}[2]{*}{HV} & MOP   & 5.57  & 5.57  & 4.00  & 3.00  & 2.00  & 6.71  & 8.00  & \cellcolor[rgb]{0.851, 0.851, 0.851} 0.57 & \cellcolor[rgb]{0.851, 0.851, 0.851} \textbf{0.29} \\
		& UF    & 4.00  & 2.40  & 5.50  & 3.30  & 3.90  & 5.30  & 5.80  & \cellcolor[rgb]{0.851, 0.851, 0.851} \textbf{1.90} & \cellcolor[rgb]{0.851, 0.851, 0.851} 2.40 \\
		& WFG   & 3.67  & 4.00  & 6.89  & 2.67  & 5.00  & 3.78  & 7.89  & \cellcolor[rgb]{0.851, 0.851, 0.851} \textbf{0.56} & \cellcolor[rgb]{0.851, 0.851, 0.851} 0.89 \\
		& Overall & 4.31  & 3.81  & 5.58  & 3.00  & 3.77  & 5.15  & 7.12  & \cellcolor[rgb]{0.851, 0.851, 0.851} \textbf{1.08} & \cellcolor[rgb]{0.851, 0.851, 0.851} 1.31 \\
		\bottomrule
	\end{tabular}%
\end{table*}

\begin{table*}[!t]
	\scriptsize
	\caption{Average Performance Scores on 3-, 5-, 8- and 10-Objective WFG Test Instances.}
	\label{tab:score-many}
	\centering
	% Table generated by Excel2LaTeX from sheet 'v4_final'
	\begin{tabular}{ccccccccc}
		\toprule
		Metric & Problem & ~~DRA~~   & MOEA/DD & PICEA-g & NSGA-III & ~~HypE~~  & AOOSTM & AMOSTM \\
		\midrule
		\multirow{5}[2]{*}{HV} & $m=3$   & 4.78  & 3.67  & 2.33  & 1.89  & 5.89  & \cellcolor[rgb]{0.851, 0.851, 0.851} \textbf{0.56} & \cellcolor[rgb]{0.851, 0.851, 0.851} 0.67 \\
		& $m=5$   & 5.00  & 3.56  & 2.00  & 3.00  & 5.67  & \cellcolor[rgb]{0.851, 0.851, 0.851} \textbf{0.11} & \cellcolor[rgb]{0.851, 0.851, 0.851} 0.89 \\
		& $m=8$   & 5.22  & 3.56  & 2.89  & 2.22  & 4.78  & \cellcolor[rgb]{0.851, 0.851, 0.851} \textbf{1.44} & \cellcolor[rgb]{0.851, 0.851, 0.851} 1.78 \\
		& $m=10$  & 5.78  & 3.56  & 2.56  & 2.11  & 4.78  & \cellcolor[rgb]{0.851, 0.851, 0.851} \textbf{1.00} & \cellcolor[rgb]{0.851, 0.851, 0.851} 1.67 \\
		& Overall & 5.19  & 3.58  & 2.44  & 2.31  & 5.28  & \cellcolor[rgb]{0.851, 0.851, 0.851} \textbf{0.78} & \cellcolor[rgb]{0.851, 0.851, 0.851} 1.25 \\
		\bottomrule
	\end{tabular}%
\end{table*}

We summarize the total ranks of different algorithms based on the mean metric values in \pref{tab:rank-multi} and \pref{tab:rank-many} and compute their final ranks over all test instances. As can be seen from the tables, the proposed MOEA/D-AOOSTM and MOEA/D-AMOSTM remain the two best algorithms on all test instances. It seems that MOEA/D-AOOSTM achieves better performance under the HV metric while MOEA/D-AMOSTM obtains better IGD results.

To analyze the relative performance of different algorithms in a statistical point of view, we adopt the performance score~\cite{BaderZ11} to qualify the algorithms. Given $K$ algorithms $\{A_1,\cdots,A_K\}$, the performance score of each algorithm $A_i$, $i\in\{1,\cdots,K\}$ on a certain test instance, is defined as:
\begin{equation}
	P(A_i)=\sum_{j=1,j\neq i}^{K}\delta_{i,j},
\end{equation}
where $\delta_{i,j}=1$ when $A_i$ is significantly outperformed by $A_j$; otherwise, $\delta_{i,j}=0$. This time, we use the Kolmogorov-Smirnov test to examine whether an algorithm is significantly outperformed by another. The performance score indicates how many other algorithms are significantly better than the corresponding algorithm on a certain test instance. Thus, the smaller the performance score, the better the algorithm. \pref{tab:score-multi} and \pref{tab:score-many} present the average performance scores of different algorithms over all test instances, where we add a gray background to the top two algorithms and highlight the best algorithms in boldface. Coincident with the total ranks on mean metric values, MOEA/D-AOOSTM and MOEA/D-AMOSTM obtain the best average performance scores on all test instances with different number of objectives. Comparing with all other algorithms, the leads of the two proposed algorithms employing stable matching-based selection with incomplete lists are statistically significant.

% !Tex root = main.tex

\section{Conclusions}
\label{sec:conclusion}

The stable matching-based selection mechanism paves the way to address the balance between convergence and diversity from the perspective of achieving the equilibrium between the preferences of subproblems and solutions. However, considering the population diversity, it might not be appropriate to allow each solution to match with any subproblem on its preference list. This paper has introduced the incomplete preference lists into the stable matching-based selection model, in which the length of the preference list of each solution is restricted so that a solution is only allowed to match with one of its favorite subproblems. To overcome the drawbacks of incomplete preference lists, a two-level one-one stable matching-based selection mechanism and a many-one stable matching-based selection mechanism are proposed and integrated into MOEA/D. In particular, the length of the preference list of each solution is problem dependent and is related to the difficulty of the corresponding subproblem. To address this issue, an adaptive mechanism is proposed to dynamically control the length of the preference list of each solution according to the local competitiveness information. Comprehensive experiments are conducted on 62 benchmark problems which cover various characteristics, e.g., multi-modality, deceptive, complicated PSs and many objectives. From the experimental studies in \pref{sec:experiments}, we can clearly observe the competitive performance obtained by our proposed MOEA/D-AOOSTM and MOEA/D-AMOSTM, comparing with a variety of state-of-the-art EMO algorithms.

Although our proposed MOEA/D-AOOSTM and MOEA/D-AMOSTM have shown very competitive performance in the empirical studies, we also notice that the stable matching relationship between subproblems and solutions may sacrifice the convergence property of the population to some extent. One possible reason might be both the two-level one-one matching and the many-one matching restricts that each solution can only be selected by at most one subproblem. Future work could be focused on assigning higher priorities for elite solutions to produce offspring solutions or allowing elite solutions to be matched with more than one subproblem. It is also interesting to apply the proposed algorithms to real-world application scenarios.
%The adaptive two-level stable matching-based selection improves the diversity ability of the original stable matching-based selection. However, one issue we found in the experimental studies is that both versions of stable matching result in a slight drawback on the convergence ability of MOEA/D. The reason might be that each solution can be selected by at most one subproblem during the environmental selection. Future work could be focused on assigning higher priorities for elite solutions to produce offspring solutions or allowing each elite solutions to be matched to more than one subproblem. Another future research topic could be studies on the performance of stable matching-based selection for many-objective optimization problems.

\bibliographystyle{IEEEtran}
\bibliography{IEEEabrv,ASTM}

\end{document}

\title{Supplemental File of ``Matching-Based Selection with Incomplete Lists for Decomposition Multi-Objective Optimization''}

\author{Mengyuan Wu,
	Ke Li,
	Sam Kwong,~\IEEEmembership{Fellow,~IEEE},
	Yu Zhou and
	Qingfu Zhang,~\IEEEmembership{Fellow,~IEEE}\\
	% \textbf{COIN Report Number 2014015}
	\thanks{This work was partially supported by Hong Kong RGC General Research Fund GRF grant 9042038 (CityU 11205314).}
	\thanks{M. Wu, S. Kwong, Y. Zhou and Q. Zhang are with the Department of Computer Science, City University of Hong Kong, Kowloon, Hong Kong SAR (e-mail: mengyuan.wu@my.cityu.edu.hk, cssamk@cityu.edu.hk, yzhou57-c@my.cityu.edu.hk, qingfu.zhang@cityu.edu.hk)}
	\thanks{K. Li is with the College of Engineering, Mathematics and Physical Sciences, University of Exeter, Exeter, EX4 4QF, UK (e-mail: k.li@exeter.ac.uk)}}

\markboth{} {Wu\MakeLowercase{\textit{et al.}}: Matching-Based Selection with Incomplete Lists for Decomposition Multi-Objective Optimization}

\maketitle

\begin{abstract}
	The paper ``Matching-Based Selection with Incomplete Lists for Decomposition Multi-Objective Optimization'' introduces the concept of incomplete preference lists into the stable matching model to remedy the loss of population diversity. In particular, each solution is only allowed to maintain a partial preference list consisting of its favorite subproblems. We implement two versions of stable matching-based selection mechanisms with incomplete preference lists: one achieves a two-level one-one matching and the other obtains a many-one matching. Furthermore, an adaptive mechanism is developed to automatically set the length of the incomplete preference list for each solution according to its local competitiveness. Due to the page limit of the paper, we present the plots of the final solution sets with the best IGD values on all UF, MOP and Bi-Objective WFG test instances in this supplemental file. 
\end{abstract}

% Note that keywords are not normally used for peer review papers.
\begin{IEEEkeywords}
	Multiobjective optimization, stable matching, decomposition, selection, diversity, adaptive scheme.
\end{IEEEkeywords}

% \fi
\IEEEpeerreviewmaketitle

\section{Figures}

\begin{figure*}[htbp]
	\centering
	
	\pgfplotsset{
		every axis/.append style={
			font = \Large,
			grid = major,
			thick,
			line width = 1pt,
			label style={font=\small},
			tick label style={font=\small},
			title style={font=\normalsize },
			tick style = {line width = 0.8pt, font = \Large}}}
	
	\subfloat{
		\resizebox{0.3333\textwidth}{!}{

		}
	}
	\caption{Final solution sets with best IGD metric values found by 9 MOEAs on MOP1.}
	\label{fig:mop1_moea}
\end{figure*}

\begin{figure*}[htbp]
	\centering
	
	\pgfplotsset{
		every axis/.append style={
			font = \Large,
			grid = major,
			thick,
			line width = 1pt,
			label style={font=\small},
			tick label style={font=\small},
			title style={font=\normalsize },
			tick style = {line width = 0.8pt, font = \Large}}}
	
	\subfloat{
		\resizebox{0.3333\textwidth}{!}{
			%
		}
	}
	\caption{Final solution sets with best IGD metric values found by 9 MOEAs on MOP2.}
	\label{fig:mop2_moea}
\end{figure*}

\begin{figure*}[htbp]
	\centering
	
	\pgfplotsset{
		every axis/.append style={
			font = \Large,
			grid = major,
			thick,
			line width = 1pt,
			label style={font=\small},
			tick label style={font=\small},
			title style={font=\normalsize },
			tick style = {line width = 0.8pt, font = \Large}}}
	
	\subfloat{
		\resizebox{0.3333\textwidth}{!}{
			%
		}
	}
	\caption{Final solution sets with best IGD metric values found by 9 MOEAs on MOP3.}
	\label{fig:mop3_moea}
\end{figure*}

\begin{figure*}[htbp]
	\centering
	
	\pgfplotsset{
		every axis/.append style={
			font = \Large,
			grid = major,
			thick,
			line width = 1pt,
			label style={font=\small},
			tick label style={font=\small},
			title style={font=\normalsize },
			tick style = {line width = 0.8pt, font = \Large}}}
	
	\subfloat{
		\resizebox{0.3333\textwidth}{!}{
			%
		}
	}
	\caption{Final solution sets with best IGD metric values found by 9 MOEAs on MOP7.}
	\label{fig:mop7_moea}
\end{figure*}

\clearpage

\begin{figure*}[htbp]
	\centering
	
	\pgfplotsset{
		every axis/.append style={
			font = \Large,
			grid = major,
			thick,
			line width = 1pt,
			label style={font=\small},
			tick label style={font=\small},
			title style={font=\normalsize },
			tick style = {line width = 0.8pt, font = \Large}}}
	
	\subfloat{
		\resizebox{0.3333\textwidth}{!}{
			%
		}
	}
	\caption{Final solution sets with best IGD metric values found by 9 MOEAs on UF1.}
	\label{fig:uf1_moea}
\end{figure*}

\begin{figure*}[htbp]
	\centering
	
	\pgfplotsset{
		every axis/.append style={
			font = \Large,
			grid = major,
			thick,
			line width = 1pt,
			label style={font=\small},
			tick label style={font=\small},
			title style={font=\normalsize },
			tick style = {line width = 0.8pt, font = \Large}}}
	
	\subfloat{
		\resizebox{0.3333\textwidth}{!}{
			%
		}
	}
	\caption{Final solution sets with best IGD metric values found by 9 MOEAs on UF2.}
	\label{fig:uf2_moea}
\end{figure*}

\begin{figure*}[htbp]
	\centering
	
	\pgfplotsset{
		every axis/.append style={
			font = \Large,
			grid = major,
			thick,
			line width = 1pt,
			label style={font=\small},
			tick label style={font=\small},
			title style={font=\normalsize },
			tick style = {line width = 0.8pt, font = \Large}}}
	
	\subfloat{
		\resizebox{0.3333\textwidth}{!}{
			%
		}
	}
	\caption{Final solution sets with best IGD metric values found by 9 MOEAs on UF3.}
	\label{fig:uf3_moea}
\end{figure*}

\begin{figure*}[htbp]
	\centering
	
	\pgfplotsset{
		every axis/.append style={
			font = \Large,
			grid = major,
			thick,
			line width = 1pt,
			label style={font=\small},
			tick label style={font=\small},
			title style={font=\normalsize },
			tick style = {line width = 0.8pt, font = \Large}}}
	
	\subfloat{
		\resizebox{0.3333\textwidth}{!}{
			%
		}
	}
	\caption{Final solution sets with best IGD metric values found by 9 MOEAs on UF10.}
	\label{fig:uf10_moea}
\end{figure*}

\clearpage

\begin{figure*}[htbp]
	\centering
	
	\pgfplotsset{
		every axis/.append style={
			font = \Large,
			grid = major,
			thick,
			line width = 1pt,
			label style={font=\small},
			tick label style={font=\small},
			title style={font=\normalsize },
			tick style = {line width = 0.8pt, font = \Large}}}
	
	\subfloat{
		\resizebox{0.3333\textwidth}{!}{
			%
		}
	}
	\caption{Final solution sets with best IGD metric values found by 9 MOEAs on WFG1.}
	\label{fig:wfg1_moea}
\end{figure*}

\begin{figure*}[htbp]
	\centering
	
	\pgfplotsset{
		every axis/.append style={
			font = \Large,
			grid = major,
			thick,
			line width = 1pt,
			label style={font=\small},
			tick label style={font=\small},
			title style={font=\normalsize },
			tick style = {line width = 0.8pt, font = \Large}}}
	
	\subfloat{
		\resizebox{0.3333\textwidth}{!}{
			%
		}
	}
	\caption{Final solution sets with best IGD metric values found by 9 MOEAs on WFG2.}
	\label{fig:wfg2_moea}
\end{figure*}

\begin{figure*}[htbp]
	\centering
	
	\pgfplotsset{
		every axis/.append style={
			font = \Large,
			grid = major,
			thick,
			line width = 1pt,
			label style={font=\small},
			tick label style={font=\small},
			title style={font=\normalsize },
			tick style = {line width = 0.8pt, font = \Large}}}
	
	\subfloat{
		\resizebox{0.3333\textwidth}{!}{
			%
		}
	}
	\caption{Final solution sets with best IGD metric values found by 9 MOEAs on WFG3.}
	\label{fig:wfg3_moea}
\end{figure*}

\begin{figure*}[htbp]
	\centering
	
	\pgfplotsset{
		every axis/.append style={
			font = \Large,
			grid = major,
			thick,
			line width = 1pt,
			label style={font=\small},
			tick label style={font=\small},
			title style={font=\normalsize },
			tick style = {line width = 0.8pt, font = \Large}}}
	
	\subfloat{
		\resizebox{0.3333\textwidth}{!}{
			%
		}
	}
	\caption{Final solution sets with best IGD metric values found by 9 MOEAs on WFG9.}
	\label{fig:wfg9_moea}
\end{figure*}

\bibliographystyle{IEEEtran}
\bibliography{IEEEabrv,ASTM}